%% file: main.tex
\PassOptionsToPackage{table}{xcolor}
\documentclass{article}

\usepackage{microtype}
\usepackage{graphicx}
\usepackage{booktabs}
\usepackage{hyperref}
\usepackage[preprint]{preprint}

\usepackage{amsmath}
\usepackage{amssymb}
\usepackage{mathtools}
\usepackage{amsthm}
\usepackage{xcolor}
\usepackage{enumitem}
\usepackage{multirow}
\usepackage{makecell}
\usepackage{diagbox}
\usepackage{adjustbox}  
\usepackage{colortbl}
\usepackage[ruled,linesnumbered,vlined]{algorithm2e}  

\theoremstyle{plain}
\newtheorem{theorem}{Theorem}[section]
\newtheorem{proposition}[theorem]{Proposition}
\newtheorem{lemma}[theorem]{Lemma}

\definecolor{high1}{RGB}{172, 226, 225}   
\definecolor{high}{RGB}{218, 255, 251}    
\definecolor{gap}{RGB}{185, 229, 232}     

\newcommand{\none}{\cellcolor{white}}

\newcommand{\tbd}{\textcolor{red}{--}}

\newcommand{\dthree}{D$^3$}

\newcommand{\rdms}{RDS-MOPD}

\pptitlerunning{Robust Domain-Mixture Scheduling for Multi-Teacher On-Policy Distillation}

\begin{document}

\twocolumn[
  \pptitle{Robust Domain-Mixture Scheduling for\\Multi-Teacher On-Policy Distillation}

  \begin{ppauthorlist}
    \ppauthor{Yankai Chen}{mcgill,mbzuai}
    \ppauthor{Sida Lin}{hkust}
    \ppauthor{Yifan Li}{cuhk}
    \ppauthor{Hanjun Luo}{nyu}
    \ppauthor{Hanrong Zhang}{uic}
    \ppauthor{Zhuohan Xie}{mbzuai}
    \ppauthor{Bowei He}{mcgill,mbzuai}
    \ppauthor{Wei-Chieh Huang}{uic}
    \ppauthor{Meng Ding}{umb}
    \ppauthor{Wei Xue}{hkust}
    \ppauthor{Yike Guo}{hkust}
    \ppauthor{Philip S. Yu}{uic}
    \ppauthor{Xue Liu}{mcgill,mbzuai}
  \end{ppauthorlist}

  \ppaffiliation{mcgill}{McGill University, Montreal, Canada}
  \ppaffiliation{mbzuai}{Mohamed bin Zayed University of Artificial Intelligence, Abu Dhabi, UAE}
  \ppaffiliation{hkust}{The Hong Kong University of Science and Technology, Hong Kong, China}
  \ppaffiliation{cuhk}{The Chinese University of Hong Kong, Hong Kong, China}
  \ppaffiliation{nyu}{New York University, New York, USA}
  \ppaffiliation{uic}{University of Illinois Chicago, Chicago, USA}
  \ppaffiliation{umb}{University of Massachusetts Boston, Boston, USA}

  \ppcorrespondingauthor{Yankai Chen}{yankaichen@acm.org}

  \ppkeywords{on-policy distillation, multi-teacher distillation, data mixture scheduling, distributionally robust optimization, large language models}

  \vskip 0.3in
]

\printAffiliationsAndNotice{}

\input{sections/0_abstract}

\input{sections/1_introduction}
\input{sections/2_background}
\input{sections/3_method}
\input{sections/4_experiments}

\input{sections/7_related}
\input{sections/8_conclusion}

\bibliographystyle{preprint}
\bibliography{refs}

\input{sections/9_appendix}
\input{sections/9b_proofs}

\end{document}

%% file: sections/0_abstract.tex
\begin{abstract}
Multi-teacher on-policy distillation (MOPD) merges the capabilities of multiple domain-expert teachers,
such as mathematical reasoning, code generation, and instruction following, into a single compact student.
Since all domains share a limited training budget, how to allocate it across domains directly determines
the quality of distillation.
Existing methods allocate the budget by each domain's learning progress, but offer no guarantee for lagging
domains.
We therefore introduce distributionally robust optimization (DRO) into MOPD to prioritize the worst-performing
domain.
However, the per-domain training signals are estimated from a few rollouts and are inherently noisy,
so directly applying DRO mistakes noise for the worst case, lets occasional spikes hijack the budget,
and protects inherently hard domains rather than truly regressing ones.
To address these issues, we propose \rdms{}, which keeps the DRO objective and redesigns its three parts:
it shifts the allocation only when differences clearly exceed the noise, bounds the influence of any single
anomaly, and defines the worst case as the regression from each domain's own best level,
all with theoretical guarantees.
Experiments show that \rdms{} outperforms recent competing methods and ranks first on most
mathematical reasoning, code generation, and instruction-following benchmarks.
\end{abstract}

%% file: sections/1_introduction.tex
\section{Introduction}
\label{sec:intro}

As large language model (LLM) services become deeply embedded in various real-world applications,
users often issue requests of vastly different types within one continuous session,
and expect the same model to both handle every task well and keep the conversation
coherent~\citep{DBLP:conf/iclr/ZhaoRHCCD24}.
Deploying a dedicated model for each type of task multiplies the inference and maintenance cost,
and breaks the coherence of the session whenever the task switches.
Consequently, merging multiple expert capabilities into a single compact student model has become a
fundamental requirement for large-scale
deployment~\citep{DBLP:journals/corr/HintonVD15}.
\textit{Multi-teacher on-policy distillation} (MOPD) is a promising post-training paradigm toward
this goal~\citep{DBLP:journals/corr/abs-2606-30406,DBLP:journals/corr/abs-2608-19098,DBLP:journals/corr/abs-2608-24987},
where the student learns from multiple domain-expert teachers simultaneously on its own sampled
responses.
Compared to offline sequence-level distillation,
on-policy distillation provides dense supervision directly on the student's own distribution,
which alleviates the distribution shift between training and
inference~\citep{DBLP:conf/iclr/AgarwalVZSGGB24,DBLP:conf/iclr/GuDWH24}.

In MOPD, all domains (e.g., mathematics, code,
and instruction following (IF)) share a limited training budget.
Hence, the \textit{domain mixture}, i.e.,
how much the student learns from each teacher at every step,
directly determines the quality of distillation.
Open-MOPD~\citep{DBLP:journals/corr/abs-2608-19098} diagnoses the capability imbalance in multi-teacher distillation,
and proposes to balance the token shares and follow the per-domain KL gaps.
More recently, D$^3$-MOPD~\citep{DBLP:journals/corr/abs-2608-24987} dynamically allocates the budget according to the
\textit{learning progress} of each domain (i.e., the remaining KL gap and its descent velocity),
steering the budget toward the domains that benefit the most.
However, these methods all aim to accelerate the overall progress: they only favor ``which domain improves the fastest'' and offer no guarantee for lagging or regressing domains.
In multi-capability merging, what users perceive is often the weakest capability,
and the lag of any single capability directly hurts the service quality.

\begin{figure}[t]
    \centering
    \includegraphics[width=\linewidth]{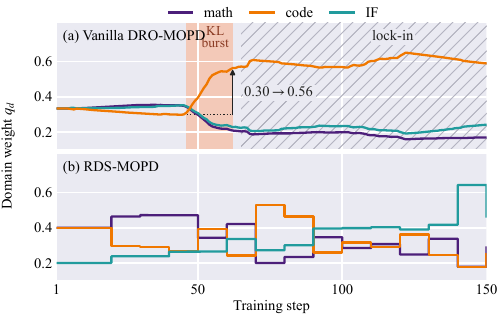}
    \caption{Domain weights on the training trajectory. (a) Vanilla DRO-MOPD: a code KL burst (shaded) lifts the code weight from 0.30 to 0.56, which then locks in (hatched); (b) \rdms{} keeps re-allocating the budget without lock-in.}
    \label{fig:teaser}
\end{figure}

To this end, we introduce \textit{distributionally robust optimization} (DRO) into
domain-mixture scheduling for MOPD.
DRO makes decisions against the most unfavorable case within an uncertainty set,
and thereby prioritizes the worst-performing group~\citep{DBLP:conf/iclr/SagawaKHL20,DBLP:journals/mor/Lam16}.
This idea naturally matches the goal of ``leaving no capability behind''.
Different from prior work that determines the pre-training data mixture offline with a proxy
model~\citep{DBLP:conf/nips/XiePDLLLLLY23}, to the best of our knowledge, this is the first work that applies DRO
to online mixture scheduling in MOPD.
Nevertheless, directly applying DRO to MOPD (referred to as
\textit{Vanilla DRO-MOPD}) is confronted with a fundamental issue:
every decision of DRO relies on the estimated per-domain training signals,
which are computed from a limited number of rollouts per domain and are inherently noisy.
As shown in Figure~\ref{fig:teaser}(a),
on the training trajectory of our Vanilla DRO-MOPD implementation,
the weight of the code domain surges after a KL burst and stays high until the end of training,
even though code is no longer the worst-performing domain.
Therefore, it is non-trivial to make reliable robust decisions under noisy training signals,
and Vanilla DRO-MOPD mainly suffers from the following three instabilities (diagnosed in \S\ref{sec:method-dro}).
\begin{itemize}[leftmargin=*]
    \item \textbf{Worst-Case Misjudgment.} Vanilla DRO-MOPD moves its weights at the same rate
        whether the evidence is strong or weak,
        so it systematically up-weights the domain whose noisy estimate happens to be high.
        On the above trajectory, the current worst domain changes 52 times within 145 training
        steps, and on more than half (54\%) of the non-spike steps,
        the gap between the worst and runner-up domains is below one noise standard
        deviation.
    \item \textbf{Spike Hijacking.} Every observation enters the weights of Vanilla DRO-MOPD in
        proportion to its magnitude and is carried forward by all later updates,
        so an abnormal KL spike can seize a large share of the budget.
        On the above trajectory, the KL of the code domain rises to about 30$\times$ its typical
        level within a dozen steps, and its weight nearly doubles
        and stays high until the end.
    \item \textbf{Misplaced Protection.} DRO is intended to protect the weak domains,
        but it compares the magnitude of losses,
        while the KL values given by different teachers naturally differ in scale.
        As a result, the budget flows to the domains with inherently large KL rather than the truly
        regressing ones: with the raw KL as the loss,
        Vanilla DRO-MOPD eventually allocates the whole budget to the code domain,
        whereas the IF domain, which regresses the most from its own best level,
        receives almost no protection.
\end{itemize}

To tackle these challenges, we propose a novel framework,
namely \textit{\underline{R}obust \underline{D}omain-Mixture \underline{S}cheduling for
\underline{M}ulti-Teacher \underline{O}n-\underline{P}olicy \underline{D}istillation} (\rdms{}).
Our key observation is that the three instabilities do not stem from the idea of DRO itself;
instead, Vanilla DRO-MOPD builds all three of its parts (i.e., the deviation radius,
the nominal allocation, and the definition of the ``worst case'') upon unreliable signals.
Hence, \rdms{} keeps the DRO objective and redesigns its three parts with three modules.
Specifically, (1) to handle worst-case misjudgment, we propose \textbf{\underline{N}oise-aware \underline{R}adius \underline{C}alibration} (\texttt{NRC}), which calibrates the deviation radius to the noise level, so that the evidence strength decides how far the allocation may deviate.
It lower-bounds the temperature of the nominal allocation via the online-estimated noise floor,
so that differences within the noise range only cause bounded share changes,
while truly significant ones still receive a decisive response.
(2) To handle spike hijacking, we design \textbf{\underline{S}pike-robust \underline{A}llocation \underline{E}stimation} (\texttt{SAE}), which rebuilds the nominal allocation from KL observations clipped by a log-domain Huber rule,
bounding the influence of spikes while still following persistent real changes.
Meanwhile, \texttt{SAE} re-estimates the nominal allocation at every update rather than
accumulating it from previous decisions, so an anomaly no longer persists in the budget.
(3) To handle misplaced protection, we propose \textbf{\underline{W}orst-case \underline{D}omain \underline{R}eweighting} (\texttt{WDR}), which redefines the ``worst case'' as each domain's regression from its own best level,
and exactly solves DRO around the progress-driven nominal allocation.
It thus reweights the domains toward the truly regressing ones with a bounded amount of extra budget.
As shown in Figure~\ref{fig:teaser}(b), under the same setting, \rdms{} keeps re-allocating the budget throughout training, and no domain is locked in.
Theoretically, we prove that \texttt{NRC} and \texttt{SAE} guarantee bounded share changes and
bounded influence of anomalies, respectively, and that \texttt{WDR} exactly solves a KL-ball DRO problem.

\begin{figure*}[t]
  \centering
  \includegraphics[width=\textwidth]{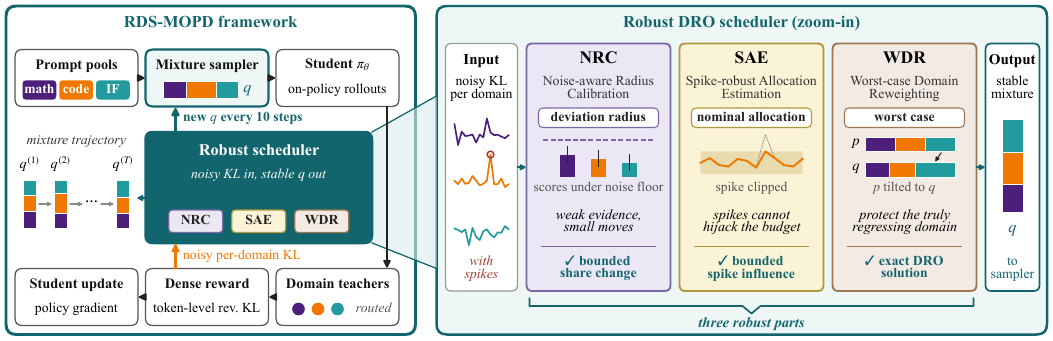}
  \caption{Illustration of our \rdms{} framework (left) and its robust scheduler (right).}
  \label{fig:framework}
\end{figure*}

We distill a 3B student model from three domain-expert teachers for mathematics, code,
and instruction following, and conduct extensive experiments on six benchmarks.
The results show that \rdms{} outperforms recent competing methods,
including Open-MOPD~\citep{DBLP:journals/corr/abs-2608-19098} and D$^3$-MOPD~\citep{DBLP:journals/corr/abs-2608-24987},
and ranks first on the majority of benchmarks.
\rdms{} also makes no large share reversal in training, and in a stress test that injects noise into the code
teacher's signal, it still keeps a stable mixture and protects the learning progress of the other domains.
These results demonstrate both the superiority of \rdms{} and the effectiveness of its modules.
To summarize, our main contributions are threefold:
\begin{itemize}[leftmargin=*]
    \item To the best of our knowledge, we are the first to introduce DRO into domain-mixture
        scheduling for MOPD, and formally diagnose three instabilities of Vanilla DRO-MOPD
        under noise.
    \item We propose \rdms{}, a single DRO structure whose three modules fix these instabilities
        with theoretical guarantees.
    \item Extensive experiments show that \rdms{} ranks first on most benchmarks,
        and validate the effectiveness of its modules.
\end{itemize}

%% file: sections/2_background.tex
\section{Preliminaries}
\label{sec:background}

We introduce MOPD and its domain-mixture scheduling problem.
Bold lowercase letters denote vectors (e.g., $\mathbf{q}$) and calligraphic letters denote sets (e.g., $\mathcal{D}$).
$\Delta^{K}$ denotes the probability simplex over $K$ domains, $\mathbf{u}=\mathbf{1}/K$ the uniform allocation
($\mathbf{1}$ is the all-ones vector), and $\mathrm{KL}(\cdot\,\|\,\cdot)$ the Kullback--Leibler divergence.
We index training steps by $s$ and scheduling updates by $t$ (omitted when clear from the context),
and summarize the notations in Table~\ref{tab:notation} of Appendix~\ref{app:notation}.

\subsection{Multi-Teacher On-Policy Distillation}
\label{sec:pre-mopd}

Let $\mathcal{D}$ be the set of domains (here, $\mathcal{D}=\{\text{math},\text{code},\text{IF}\}$) with $K=|\mathcal{D}|$.
Each domain $d$ has a prompt pool $\mathcal{P}_d$ and an RL-trained expert teacher $\pi_{\phi_d}$.
MOPD distills all teachers into one student policy $\pi_\theta$, and routes each prompt $x$ to the teacher
of its domain $d(x)$.
Given $x$, the student samples a response $y\sim\pi_\theta(\cdot\mid x)$ on-policy,
and the routed teacher scores it with a single prefill.
At each position $j$ of $y$, the teacher computes the reverse KL $D_j$ from the next-token distribution of
the student to its own, restricted to the student's top-$k$ tokens~\citep{DBLP:journals/corr/abs-2608-19098}.
The dense token-level reward $-D_j$ serves as the advantage of a critic-free policy-gradient
update~\citep{DBLP:journals/corr/SchulmanWDRK17}.
The distillation loss $\mathcal{L}_d(\theta)$ is the expected per-token $D_j$ over the domain-$d$ prompts
and their on-policy responses.
At training step $s$, the \textit{per-domain KL signal} is computed as:
\begin{equation}
  m_d^{(s)}=\frac{1}{|\mathcal{B}_d|}\sum_{y\in\mathcal{B}_d}\frac{1}{|y|}
  \sum_{j=1}^{|y|}D_j,
  \label{eq:kl-signal}
\end{equation}
where $\mathcal{B}_d$ holds the domain-$d$ responses of step $s$, and $|y|$ is the length of $y$.
As a one-batch estimate of $\mathcal{L}_d$, $m_d^{(s)}$ reuses the rewards already computed by the trainer
(no extra forward pass), but is noisy, as each batch holds a limited number of rollouts per domain.

\subsection{Domain-Mixture Scheduling}
\label{sec:pre-sched}

Every few training steps, a scheduler updates the \textit{domain mixture} $\mathbf{q}^{(t)}\in\Delta^{K}$
from the observed KL signals.
Each batch then samples prompts in proportion to $\mathbf{q}^{(t)}$, with a random
jitter (Appendix~\ref{app:impl}).
Between two scheduling updates, the student thus minimizes $\sum_{d}q_d\,\mathcal{L}_d(\theta)$;
$\mathbf{q}$ only decides how many prompts each domain receives, and the per-token loss is unchanged.
Some schedulers fix the mixture before training, e.g., by prompt-pool size~\citep{DBLP:journals/corr/abs-2606-30406}, or determine it offline with a proxy
model~\citep{DBLP:conf/nips/XiePDLLLLLY23}.
Others adapt to each domain's learning progress online, e.g., Open-MOPD~\citep{DBLP:journals/corr/abs-2608-19098}
reweights the per-domain rewards by token-share balancing and KL-gap following, and
\dthree{}-MOPD~\citep{DBLP:journals/corr/abs-2608-24987} resamples domains by the remaining KL gap times its descent velocity.

%% file: sections/3_method.tex
\section{Our \rdms{} Methodology}
\label{sec:method}

We first formulate DRO for MOPD and diagnose Vanilla DRO-MOPD (\S\ref{sec:method-dro}), then present \rdms{}
with its three modules (\S\ref{sec:method-structure}, Figure~\ref{fig:framework}) and its overall procedure
(\S\ref{sec:method-alg}).

\subsection{DRO for MOPD}
\label{sec:method-dro}

\subsubsection{\textbf{DRO Formulation for MOPD}}
\label{sec:method-dro-form}
DRO decides against the most unfavorable distribution within an uncertainty set around a
nominal one~\citep{DBLP:journals/mansci/Ben-TalHWMR13,DBLP:journals/aos/DuchiN21}.
In MOPD, where the student minimizes $\sum_d q_d\mathcal{L}_d$ (\S\ref{sec:pre-sched}), a DRO scheduler
picks the most unfavorable mixture, up-weighting domains with a higher estimated loss.
With a KL penalty around a nominal allocation $\mathbf{p}$, the DRO decision is:
\begin{equation}
  \mathbf{q}=\mathop{\arg\max}_{\mathbf{q}'\in\mathcal{U}}\
  \Big\{\big\langle\mathbf{q}',\hat{\boldsymbol{\ell}}\big\rangle
  -\frac{1}{\eta}\,\mathrm{KL}\big(\mathbf{q}'\,\|\,\mathbf{p}\big)\Big\},
  \label{eq:dro}
\end{equation}
where $\mathcal{U}\subseteq\Delta^{K}$ is the feasible set.
Eqn.~\eqref{eq:dro} has three parts:
(i) the \textit{deviation radius}, set by $\eta$ (a larger $\eta$ lets $\mathbf{q}$ move farther);
(ii) the \textit{nominal allocation} $\mathbf{p}$, the anchor that $\mathbf{q}$ moves from;
and (iii) the \textit{definition of the worst case}, i.e., the loss estimate $\hat{\boldsymbol{\ell}}$ of
$(\mathcal{L}_d)_{d\in\mathcal{D}}$.

\textit{Vanilla DRO-MOPD} applies online Group DRO~\citep{DBLP:conf/iclr/SagawaKHL20} to MOPD with each
domain as a group, where $\mathbf{q}$ may reweight either the losses or the sampling.
At every training step $s$ (so $t=s$), it solves Eqn.~\eqref{eq:dro} with $\mathcal{U}=\Delta^{K}$, whose
closed form is $q_d\propto p_d\,e^{\eta\hat\ell_d}$.
Its nominal allocation is the previous weights, $\mathbf{p}=\mathbf{q}^{(s-1)}$ with
$\mathbf{q}^{(0)}=\mathbf{u}$, and its loss is the KL signal normalized by its initial level $I_d$ (early mean of $m_d$), $\hat\ell_d^{(s)}=m_d^{(s)}/I_d$, giving the exponentiated-gradient update:
\begin{equation}
  q_d^{(s)}\ \propto\ q_d^{(s-1)}\,e^{\eta\,\hat\ell_d^{(s)}}.
  \label{eq:hedge}
\end{equation}

\subsubsection{\textbf{Insufficiency of Vanilla DRO-MOPD}}
\label{sec:method-insuff}
We model each observation as its true level times a noise factor, i.e.,
$m_d^{(s)}=\bar m_d^{(s)}e^{\xi_d^{(s)}}$,
where the zero-mean log-noise $\xi_d$ may be heavy-tailed and differ in scale across domains.
Unrolling Eqn.~\eqref{eq:hedge} from $\mathbf{q}^{(0)}=\mathbf{u}$, taking the log-ratio of two domains $d$ and $d'$
(which cancels the normalization), and writing $e^{\xi_d}=1+(e^{\xi_d}-1)$ splits it into a signal term and a
noise term summed over steps $s'\le s$:\predisplaypenalty=0
\begin{equation}
  \adjustbox{max width=\columnwidth}{$\begin{aligned}
  \log\frac{q_d^{(s)}}{q_{d'}^{(s)}}
  =\ &\underbrace{\eta\sum_{s'\le s}\Big(\frac{\bar m_d^{(s')}}{I_d}-\frac{\bar m_{d'}^{(s')}}{I_{d'}}\Big)}_{\text{signal}}\\
  &+\underbrace{\eta\sum_{s'\le s}\Big(\frac{\bar m_d^{(s')}}{I_d}\big(e^{\xi_d^{(s')}}-1\big)
  -\frac{\bar m_{d'}^{(s')}}{I_{d'}}\big(e^{\xi_{d'}^{(s')}}-1\big)\Big)}_{\text{noise}}.
  \end{aligned}$}
  \label{eq:hedge-ratio}
\end{equation}
Eqn.~\eqref{eq:hedge-ratio} shows how each part breaks under the noise.
(1) \textit{Worst-case misjudgment (deviation radius).}
The radius, fixed by $\eta$, ignores the noise level, so the noise term accumulates like a random walk whose
spread grows with the number of steps, and even drifts toward the noisier domain since $\mathbb{E}[e^{\xi}]\ge 1$.
(2) \textit{Spike hijacking (nominal allocation).}
Anchoring on the previous weights accumulates every observation, so a single observation $m_d^{(s_0)}=M$
leaves a term $\eta M/I_d$ in the log-weight ratio of domain $d$ to any other domain at all later steps,
which grows linearly in $M$ and never fades (e.g., the code burst in Figure~\ref{fig:teaser}(a)).
(3) \textit{Misplaced protection (definition of the worst case).}
Let $\bar m_d^{\star}$ be the lowest true level of domain $d$ so far; the true level then splits into three factors:
\begin{equation}
  \bar m_d=\underbrace{I_d}_{\text{teacher scale}}\cdot
  \underbrace{\frac{\bar m_d^{\star}}{I_d}}_{\text{descent}}\cdot
  \underbrace{\frac{\bar m_d}{\bar m_d^{\star}}}_{\text{regression}}.
  \label{eq:factor}
\end{equation}
Normalizing by $I_d$ removes only the teacher scale, so a domain that descended far but then regresses still
ranks below a barely descended one.
Thus, the problem lies not in the robust objective but in its three parts, calling for a noise-aware
radius, a spike-robust nominal allocation, and a worst case defined by regression alone.

\subsection{\rdms{}}
\label{sec:method-structure}
\rdms{} keeps the robust objective and redesigns its three parts.
At its $t$-th update, it solves:
\begin{equation}
  \mathbf{q}^{(t)}=\mathop{\arg\max}_{\mathbf{q}\in\mathcal{U}}\
  \Big\{\big\langle\mathbf{q},\mathbf{h}\big\rangle-\frac{1}{\eta}\,
  \mathrm{KL}\big(\mathbf{q}\,\|\,\mathbf{p}\big)\Big\},
  \label{eq:rds-dro}
\end{equation}
where $\mathbf{p}$ is the nominal allocation, $\mathbf{h}$ the excess loss,
$\eta$ the tilt strength, and $\mathcal{U}$ the feasible set,
i.e., Eqn.~\eqref{eq:dro} with $\mathbf{h}$ as the loss.

The nominal allocation is recomputed at every update from the progress score~\citep{DBLP:journals/corr/abs-2608-24987} $r_d=v_d\,\ell_d/I_d$ of each
domain, where $\ell_d$ is the exponential moving average (EMA) of the clipped KL of domain $d$
(\S\ref{sec:method-hr}), $\ell_d/I_d$ is the remaining gap, and the descent velocity $v_d\ge 0$ is the
relative decrease of $\ell_d$ over recent steps (Appendix~\ref{app:sched}):
\begin{equation}
  \mathbf{p}=f\,\mathbf{1}+(1-Kf)\,\mathrm{softmax}(\mathbf{r}/\tau),
  \label{eq:rds-nominal}
\end{equation}
where $f$ is the share floor and $\tau$ the temperature, so a domain with a large remaining gap that
still descends fast receives a larger share.
The softmax solves Eqn.~\eqref{eq:dro} with $\mathbf{r}$, $\mathbf{u}$, and $\tau$ in place of
$\hat{\boldsymbol{\ell}}$, $\mathbf{p}$, and $1/\eta$.
Hence, $\tau$ sets the radius that turns noisy scores into shares (a larger $\tau$ keeps
$\mathbf{p}$ closer to $\mathbf{u}$), while $\eta$ only bounds the step from $\mathbf{p}$ to $\mathbf{q}$.

Accordingly, three modules redesign the three parts:
(i) \texttt{NRC} calibrates the radius $\tau$ to the noise level (\S\ref{sec:method-nf});
(ii) \texttt{SAE} re-estimates $\mathbf{p}$ at every update from clipped observations instead of the previous
weights (\S\ref{sec:method-hr});
(iii) \texttt{WDR} defines the worst case as each domain's excess loss $\mathbf{h}$ over its best level
and solves Eqn.~\eqref{eq:rds-dro} exactly (\S\ref{sec:method-tilt}).

\begin{figure}[t]
  \centering
  \includegraphics[width=\linewidth]{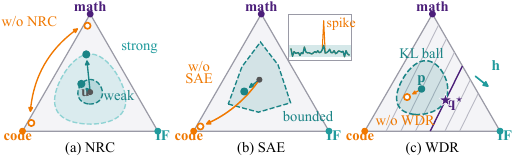}
  \caption{Three modules on the mixture simplex.
  (a) \texttt{NRC} sizes the KL ball around $\mathbf{u}$ by evidence; without it, $\mathbf{p}$ flips between vertices.
  (b) \texttt{SAE} keeps $\mathbf{p}$ bounded after a code spike; without it, $\mathbf{p}$ is dragged to the code vertex.
  (c) \texttt{WDR}'s solution $\mathbf{q}^{\star}$ lies where the highest iso-line of $\langle\mathbf{q},\mathbf{h}\rangle$ touches the KL ball of $\mathbf{p}$.}
  \label{fig:modules}
\end{figure}

\subsubsection{\textbf{Noise-aware Radius Calibration (NRC)}}
\label{sec:method-nf}
\texttt{NRC} fixes \textit{worst-case misjudgment}:
a difference that does not clearly exceed the noise should not cause a large change in the allocation.
A common choice is the max-normalized temperature $\tau=T\cdot r_{\max}$~\citep{DBLP:journals/corr/abs-2608-24987}, where $T$ is the base temperature
and $r_{\max}=\max_d r_d$ is the score of the leading domain (the leader).
The softmax input is then $(\mathbf{r}/r_{\max})/T$, so the leader always has a normalized score of $1$,
whether its lead is far beyond the noise or only a noise artifact.
Near a plateau, the velocities and thus $r_{\max}$ approach $0$, so the leader switches with the noise, and each switch moves a large share.
\texttt{NRC} therefore calibrates the radius to the uncertainty of the leading score.

\textbf{Noise-Calibrated Radius.}
We estimate the scale of the log-noise $\xi_d$ (\S\ref{sec:method-insuff}) of each domain from the recent
log-differences of its clipped observations (\S\ref{sec:method-hr}).
The delta method then carries this scale to the standard error (SE) of each score (Appendix~\ref{app:sched}).
Let $\hat\sigma$ be the SE of the leading score, and $\epsilon$ the noise-floor multiplier.
\texttt{NRC} lower-bounds $r_{\max}$ by the noise floor $\epsilon\hat\sigma$:
\begin{equation}
  \tau=T\cdot\max\big(r_{\max},\ \epsilon\,\hat\sigma\big).
  \label{eq:nf-temp}
\end{equation}
Thus, as the evidence weakens, $\tau$ stays large, and $\mathbf{p}$ moves smoothly toward $\mathbf{u}$
instead of following a noisy leader (Figure~\ref{fig:modules}(a)).

\begin{proposition}[Bounded Share Change below the Noise Level]
\label{prop:nf}
Let $\mathbf{p}$ be given by Eqns.~\eqref{eq:rds-nominal} and \eqref{eq:nf-temp} with $\epsilon>0$.
Suppose that, at two consecutive updates, all scores satisfy $0\le r_d\le\zeta\hat\sigma$ for some
$0<\zeta\le\epsilon$, with $\hat\sigma>0$ taken at the same update.
In this case, no score exceeds $\zeta$ SEs, and the noise floor is active.
Then every nominal share changes by at most $\frac{(1-Kf)\,\zeta}{2\,\epsilon\,T}$.
\end{proposition}

Proposition~\ref{prop:nf} (proved in Appendix~\ref{app:proof-nf}) guarantees that differences within the
noise level only cause bounded share changes, and the weaker the evidence, the smaller the change.

\subsubsection{\textbf{Spike-robust Allocation Estimation (SAE)}}
\label{sec:method-hr}
\texttt{SAE} fixes \textit{spike hijacking}: one abnormal observation should not dominate the allocation.
Re-estimating the nominal allocation at every update stops a spike from accumulating, but the spike still
enters every statistic built on linear averages (e.g., $I_d$, $\ell_d$, and $v_d$) in proportion to its size.
\texttt{SAE} therefore first clips every new observation into a range around the recent median.

\textbf{Log-Domain Huber Clipping.}
Since the noise is multiplicative (\S\ref{sec:method-insuff}), we clip each new observation $m_d^{(s)}$ in the
log domain, i.e., by a ratio to the median of the raw observations over the previous $H$ steps,
$\mu_d^{(s)}=\mathrm{median}\{m_d^{(s-H)},\dots,m_d^{(s-1)}\}$ (fewer in warm-up):
\begin{equation}
  \tilde m_d^{(s)}=\mathrm{clip}\big(m_d^{(s)},\ e^{-c}\mu_d^{(s)},\ e^{c}\mu_d^{(s)}\big),
  \label{eq:hr}
\end{equation}
where $\mathrm{clip}(x,a,b)$ restricts $x$ to $[a,b]$, and $c$ is the clipping threshold.
Normal fluctuations pass unchanged, and a spike keeps its direction while its size is bounded.
This is the Huber $\psi$-function~\citep{Huber1964} in the log domain, and all other statistics of \rdms{}
are computed on the clipped sequence.
Since the median uses the raw observations, a single anomaly cannot drag the center.

\begin{proposition}[Bounded Influence and Trend Preservation]
\label{prop:hr}
Let $H\ge 3$ and $c>0$.
(i) Every clipped observation lies in $[e^{-c}\mu_d^{(s)},\,e^{c}\mu_d^{(s)}]$ for $s>1$.
(ii) Replacing one raw observation $m_d^{(s_0)}$ with $s_0>H$ by any $M>0$ changes the clipped sequence only
at steps $s_0,\dots,s_0+H$, and only within a range independent of $M$.
Hence any EMA level of the clipped sequence stays in a range independent of $M$, whereas without clipping it
grows linearly in $M$.
(iii) If the observations are noise-free and jump at step $s_0$ from one constant level to another, the
clipped sequence equals the raw one from step $s_0+\lfloor H/2\rfloor+1$ on.
\end{proposition}

Hence, a single spike, however large, only has a bounded and short-lived effect, keeping $\mathbf{p}$ in a
bounded region (Figure~\ref{fig:modules}(b)), while a persistent real change still passes through after a
short delay (proofs in Appendix~\ref{app:proof-hr}).
A sustained burst is thus treated as a real change, whose effect is bounded by \texttt{WDR}
(Theorem~\ref{thm:tilt}(ii)) and fades once it ends.

\subsubsection{\textbf{Worst-case Domain Reweighting (WDR)}}
\label{sec:method-tilt}
\texttt{WDR} fixes \textit{misplaced protection}:
the worst case should be a regressing domain, not an inherently hard one.
The nominal allocation alone does not protect such a domain either: a rising KL clips its $v_d$ to $0$,
giving it the smallest share.
\texttt{WDR} thus redefines the worst case and solves the DRO problem around $\mathbf{p}$.

\textbf{Excess Loss.}
We measure whether a domain is regressing by its excess loss over its own best level:
\begin{equation}
  h_d=\min\Big(\log\frac{\ell_d}{\ell_d^{\star}},\ \log\kappa\Big)\in[0,\log\kappa],
  \label{eq:excess}
\end{equation}
where $\ell_d^{\star}$ is the lowest EMA level of domain $d$ so far (current update included) and
$\kappa\ge 1$ is the excess cap.
Here, $h_d$ is the log regression factor of Eqn.~\eqref{eq:factor} on the EMA, capped to bound the tilt in Eqn.~\eqref{eq:tilt}. It is $0$ at the best level and grows as $d$ regresses; as a ratio within one domain, it depends on neither the KL scale nor $I_d$.

\textbf{Robust Best Response in a KL Ball.}
Given $\mathbf{p}$, \texttt{WDR} solves Eqn.~\eqref{eq:rds-dro} with the feasible set
$\mathcal{U}=\{\mathbf{q}\in\Delta^{K}:q_d\ge\lambda_d\}$, where $\lambda_d=\min(p_d,\lambda_{\mathrm{lo}})$
and $\lambda_{\mathrm{lo}}$ is the coverage floor: the tilt never pushes a domain below
$\min(p_d,\lambda_{\mathrm{lo}})$, and $\mathbf{p}$ itself stays feasible.
Without the floor, the solution is the closed form of \S\ref{sec:method-dro-form} with $\hat\ell_d$
replaced by $h_d$ ($\propto p_d e^{\eta h_d}$); the floor lifts any domain below $\lambda_d$ to $\lambda_d$:
\begin{equation}
  q_d=\max\big(\lambda_d,\ \nu\,p_d\,e^{\eta h_d}\big),
  \label{eq:tilt}
\end{equation}
where $\nu>0$ is the unique normalizer with $\sum_d q_d=1$ (computed exactly in Appendix~\ref{app:sched}).
Each share is thus multiplied by $e^{\eta h_d}\in[1,\kappa^{\eta}]$: the tilt is idle when all domains
are at their best, and a more regressing domain gets more budget, up to $\kappa^{\eta}$ times.

\begin{theorem}[Exact KL-Ball DRO Solution]
\label{thm:tilt}
Let $\mathbf{p}\in\Delta^{K}$ with $\mathbf{p}>\mathbf{0}$, $\mathbf{h}\in[0,\log\kappa]^{K}$, $\eta>0$,
$\kappa\ge 1$, and $\sum_d\lambda_d<1$.
(i) \textbf{Exact KL-ball DRO solution.} Eqn.~\eqref{eq:rds-dro} has a unique maximizer, given by
Eqn.~\eqref{eq:tilt}, which also maximizes $\langle\mathbf{q}',\mathbf{h}\rangle$ over the KL ball
$\{\mathbf{q}'\in\mathcal{U}:\mathrm{KL}(\mathbf{q}'\,\|\,\mathbf{p})\le\mathrm{KL}(\mathbf{q}\,\|\,\mathbf{p})\}$ around $\mathbf{p}$.
(ii) \textbf{Bounded deviation without starvation.} $\kappa^{-\eta}\le q_d/p_d\le\kappa^{\eta}$ for all $d$,
and $q_d\ge\min(f,\lambda_{\mathrm{lo}})$ for $\mathbf{p}$ of Eqn.~\eqref{eq:rds-nominal} with $Kf\le 1$.
(iii) \textbf{Targeted protection.} If all $h_d$ are equal, $\mathbf{q}=\mathbf{p}$;
otherwise, every domain with the largest $h_d$ has $q_d\ge p_d$.
\end{theorem}

Intuitively, \texttt{WDR} is an exactly solved DRO decision (proofs in Appendix~\ref{app:proof-tilt}):
within a KL ball around $\mathbf{p}$, $\mathbf{q}$ attains the largest expected excess loss, where the highest iso-line of $\langle\mathbf{q},\mathbf{h}\rangle$ touches the ball (Figure~\ref{fig:modules}(c)).

\subsection{Overall Procedure}
\label{sec:method-alg}
\texttt{SAE} clips each new KL observation at every training step (Eqn.~\eqref{eq:hr}); at every scheduling
update, \rdms{} computes $\mathbf{r}$, $\hat\sigma$, and $\mathbf{h}$ on the clipped sequences, estimates
$\mathbf{p}$ (Eqns.~\eqref{eq:rds-nominal} and \eqref{eq:nf-temp}), and tilts it into $\mathbf{q}$
(Eqn.~\eqref{eq:tilt}; Algorithm~\ref{alg:rds} in Appendix~\ref{app:impl}).
Setting $\epsilon=0$, $c=\infty$, or $\eta=0$ (then $\mathbf{q}=\mathbf{p}$) disables \texttt{NRC}, \texttt{SAE},
or \texttt{WDR}, respectively (\S\ref{sec:exp-ablation}).
An update only reads the $K$ KL sequences already computed by the trainer (Eqn.~\eqref{eq:kl-signal})
and costs $O(K^2)$, with no extra forward pass or GPU memory (\S\ref{sec:exp-rq4}).

%% file: sections/4_experiments.tex
\section{Experiments}
\label{sec:experiments}
\newcommand{\ms}{\tbd{} $\pm$ {\footnotesize \tbd{}}}

We evaluate \rdms{} by answering the following questions:
\begin{itemize}[leftmargin=*]
  \item \textbf{RQ1}: How does \rdms{} perform on math, code, and instruction following,
    compared with MOPD methods with a fixed or adaptive domain mixture and training-free model merging?
  \item \textbf{RQ2}: How does each module of \rdms{}
    contribute to the performance, and does it resolve its corresponding instability?
  \item \textbf{RQ3}: Can \rdms{} keep a robust mixture and protect the other domains
    when one domain teacher is corrupted in the middle of training?
  \item \textbf{RQ4}: Does \rdms{} incur a large time overhead?
\end{itemize}

\begin{table*}[tp]
  \centering
  \caption{Overall performance (\%, mean $\pm$ std over 3 seeds; training-free methods and reference rows have no
  std) under the BO (top) and BB (bottom) checkpoint selection criteria.
  Bold/underline: best/second-best among the ranked methods (gray reference rows excluded);
  ``Avg. Rank'' ($\downarrow$): mean per-benchmark rank over the six benchmarks;
  ``Recovery'' ($\uparrow$): $(\text{Total}-\text{SFT})/(\text{RouteRL}-\text{SFT})$ following Open-MOPD,
  where Total is the mean of the three domain averages;
  $p$-value: one-sided Wilcoxon test of \rdms{} vs.\ each method over 18 benchmark-seed pairs,
  treating pairs as independent
  (Appendix~\ref{app:protocol}).}
  \label{tab:main}
  \small
  \begin{adjustbox}{max width=\linewidth}
  \begin{tabular}{p{2.8cm}<{\centering}|c c|c c|c c|c|c|c}
    \specialrule{.1em}{0em}{0em}
    \multirow{2}{*}{Method}
      & \multicolumn{2}{c|}{Math} & \multicolumn{2}{c|}{Code} & \multicolumn{2}{c|}{Instruction Following} & \multirow{2}{*}{\makecell{Avg.\\Rank ($\downarrow$)}} & \multirow{2}{*}{\makecell{Recovery\\(\%, $\uparrow$)}} & \multirow{2}{*}{$p$-value} \\
      & AIME24 & AIME25 & LCB v5 & LCB v6 & IFEval & IFBench & & & \\
    \hline
    \rowcolor{gray!15}
    Student-SFT & 15.10 & 15.16 & 15.99 & 15.49 & 66.82 & 18.50 & -- & 0.0 & -- \\
    \rowcolor{gray!15}
    RouteRL & 24.38 & 24.32 & 22.99 & 21.94 & 73.57 & 22.50 & -- & 100.0 & -- \\
    \hline
    \multicolumn{10}{c}{\textit{Best-Overall (BO)}} \\
    \hline
    ParamMerge-Avg & 21.20 & 21.67 & 20.00 & 19.43 & 70.52 & 20.00 & 7.00 & 60.4 & $3.8\times 10^{-6}$ \\
    ParamMerge-TA & 23.33 & 24.11 & 22.46 & 21.26 & 73.20 & 22.83 & 5.00 & 94.1 & $7.6\times 10^{-6}$ \\
    \hline
    Vanilla MOPD & 23.25 $\pm$ {\footnotesize 0.29} & 23.30 $\pm$ {\footnotesize 0.24} & 21.42 $\pm$ {\footnotesize 0.12} & 21.30 $\pm$ {\footnotesize 0.24} & 72.41 $\pm$ {\footnotesize 0.29} & 21.03 $\pm$ {\footnotesize 0.13} & 5.83 & 83.6 & $3.8\times 10^{-6}$ \\
    Open-MOPD & 24.64 $\pm$ {\footnotesize 0.14} & 24.70 $\pm$ {\footnotesize 0.24} & 22.46 $\pm$ {\footnotesize 0.06} & \underline{22.31 $\pm$ {\footnotesize 0.13}} & \underline{74.26 $\pm$ {\footnotesize 0.17}} & \textbf{23.72 $\pm$ {\footnotesize 0.09}} & 2.67 & 105.6 & $4.2\times 10^{-4}$ \\
    \dthree{}-MOPD & \underline{24.77 $\pm$ {\footnotesize 0.81}} & \underline{24.90 $\pm$ {\footnotesize 0.37}} & 22.91 $\pm$ {\footnotesize 0.74} & 22.29 $\pm$ {\footnotesize 0.46} & 74.18 $\pm$ {\footnotesize 0.26} & \underline{23.42 $\pm$ {\footnotesize 0.80}} & \underline{2.58} & \underline{106.5} & $9.7\times 10^{-4}$ \\
    Vanilla DRO-MOPD & 23.85 $\pm$ {\footnotesize 0.26} & 24.69 $\pm$ {\footnotesize 0.27} & \underline{23.37 $\pm$ {\footnotesize 0.23}} & 22.07 $\pm$ {\footnotesize 0.12} & 73.97 $\pm$ {\footnotesize 0.07} & \underline{23.42 $\pm$ {\footnotesize 0.14}} & 3.42 & 103.9 & $1.6\times 10^{-4}$ \\
    \hline
    \rowcolor{high}
    \none\textbf{\rdms{}} & \textbf{26.72 $\pm$ {\footnotesize 0.16}} & \textbf{26.09 $\pm$ {\footnotesize 0.19}} & \textbf{23.79 $\pm$ {\footnotesize 0.19}} & \textbf{23.01 $\pm$ {\footnotesize 0.07}} & \textbf{74.74 $\pm$ {\footnotesize 0.10}} & 23.08 $\pm$ {\footnotesize 0.30} & \textbf{1.50} & \textbf{118.1} & -- \\
    \hline
    \multicolumn{10}{c}{\textit{Best-per-Benchmark (BB)}} \\
    \hline
    ParamMerge-Avg & 21.20 & 21.67 & 20.00 & 19.43 & 70.52 & 20.00 & 7.00 & 60.4 & $3.8\times 10^{-6}$ \\
    ParamMerge-TA & 23.33 & 24.11 & 22.46 & 21.26 & 73.20 & 22.83 & 5.33 & 94.1 & $3.8\times 10^{-6}$ \\
    \hline
    Vanilla MOPD & 23.34 $\pm$ {\footnotesize 0.14} & 23.65 $\pm$ {\footnotesize 0.09} & 21.74 $\pm$ {\footnotesize 0.06} & 21.35 $\pm$ {\footnotesize 0.15} & 72.63 $\pm$ {\footnotesize 0.21} & 21.25 $\pm$ {\footnotesize 0.14} & 5.67 & 86.5 & $3.8\times 10^{-6}$ \\
    Open-MOPD & 24.81 $\pm$ {\footnotesize 0.03} & 24.90 $\pm$ {\footnotesize 0.06} & 22.59 $\pm$ {\footnotesize 0.07} & \underline{22.38 $\pm$ {\footnotesize 0.07}} & \underline{74.55 $\pm$ {\footnotesize 0.14}} & \textbf{23.81 $\pm$ {\footnotesize 0.17}} & \underline{2.58} & 107.8 & $9.7\times 10^{-4}$ \\
    \dthree{}-MOPD & \underline{24.95 $\pm$ {\footnotesize 0.56}} & \underline{25.11 $\pm$ {\footnotesize 0.55}} & 22.91 $\pm$ {\footnotesize 0.74} & \underline{22.38 $\pm$ {\footnotesize 0.32}} & 74.54 $\pm$ {\footnotesize 0.47} & 23.50 $\pm$ {\footnotesize 0.82} & \underline{2.58} & \underline{108.7} & $9.7\times 10^{-4}$ \\
    Vanilla DRO-MOPD & 24.17 $\pm$ {\footnotesize 0.06} & 24.86 $\pm$ {\footnotesize 0.06} & \underline{23.51 $\pm$ {\footnotesize 0.03}} & 22.19 $\pm$ {\footnotesize 0.09} & 74.03 $\pm$ {\footnotesize 0.00} & \underline{23.56 $\pm$ {\footnotesize 0.10}} & 3.33 & 106.1 & $1.6\times 10^{-4}$ \\
    \hline
    \rowcolor{high}
    \none\textbf{\rdms{}} & \textbf{26.72 $\pm$ {\footnotesize 0.16}} & \textbf{26.09 $\pm$ {\footnotesize 0.19}} & \textbf{23.95 $\pm$ {\footnotesize 0.22}} & \textbf{23.20 $\pm$ {\footnotesize 0.11}} & \textbf{74.88 $\pm$ {\footnotesize 0.03}} & 23.17 $\pm$ {\footnotesize 0.17} & \textbf{1.50} & \textbf{119.5} & -- \\
    \specialrule{.1em}{0em}{0em}
  \end{tabular}
  \end{adjustbox}
\end{table*}

\subsection{Experimental Setups}
\label{sec:exp-setup}

\subsubsection{\textbf{Benchmarks and Evaluation Metrics}}
\label{sec:exp-bench}
We evaluate all methods on six benchmarks, two per domain teacher:
\textbf{AIME24} and \textbf{AIME25}~\citep{MAA:AIME} for mathematical reasoning,
\textbf{LiveCodeBench v5} and \textbf{v6} (LCB v5/v6)~\citep{DBLP:conf/iclr/JainHGLYZWSSS25} for code generation,
and \textbf{IFEval}~\citep{DBLP:journals/corr/abs-2311-07911} and \textbf{IFBench}~\citep{DBLP:journals/corr/abs-2507-02833} for instruction following (IF).
Each domain pairs an earlier benchmark with a more recent one,
so that both in-domain fitting and generalization to newer distributions are examined.
We report avg@64 on AIME24/25, avg@10 on LCB v5/v6,
and avg@4 of the strict prompt-level accuracy on IFEval/IFBench (higher is better for all).
Benchmark statistics and evaluation configurations are given in Appendix~\ref{app:benchmarks}.

\subsubsection{\textbf{Training Data and Models}}
\label{sec:exp-data}
We build our testbed upon Open-MOPD~\citep{DBLP:journals/corr/abs-2608-19098},
to our knowledge the only recent MOPD work that releases its complete training resources.
We distill the SmolLM3-3B~\citep{HF:SmolLM3} student, initialized from the released MixSFT checkpoint
(Student-SFT), from the released RL-Math, RL-Code, and RL-IF teachers.
The prompt pool contains 86,931 prompts, i.e.,
17,917 math prompts from DAPO~\citep{DBLP:journals/corr/abs-2503-14476},
23,667 code prompts from PrimeIntellect~\citep{PrimeIntellect:SYNTHETIC1} and TACO~\citep{DBLP:journals/corr/abs-2312-14852},
and 45,347 IF prompts from Nemotron-IF-RL~\citep{DBLP:journals/corr/abs-2608-19098},
where each prompt is routed to the teacher of its domain.
Per-domain statistics are given in Appendix~\ref{app:benchmarks}.

\subsubsection{\textbf{Competing Methods}}
\label{sec:exp-baselines}
We compare \rdms{} with two groups of competing methods:
(1) state-of-the-art MOPD methods with a fixed or adaptive domain mixture, i.e.,
Vanilla MOPD~\citep{DBLP:journals/corr/abs-2606-30406},
Open-MOPD~\citep{DBLP:journals/corr/abs-2608-19098},
\dthree{}-MOPD~\citep{DBLP:journals/corr/abs-2608-24987},
and Vanilla DRO-MOPD~\citep{DBLP:conf/iclr/SagawaKHL20} (\S\ref{sec:method-dro});
and (2) training-free model merging, i.e.,
ParamMerge-Avg~\citep{DBLP:conf/icml/WortsmanIGRLMNF22}
and ParamMerge-TA~\citep{DBLP:conf/iclr/IlharcoRWSHF23}.
We also report RouteRL~\citep{DBLP:journals/corr/abs-2608-19098} as a routed-teacher reference,
which is not a single deployable model and is excluded from the ranking.
Details are given in Appendix~\ref{app:baselines}.

\subsubsection{\textbf{Implementation and Evaluation}}

All MOPD methods share the same training code and sampler, and differ mainly in the mixture rule.
They use the dense top-16 teacher reverse KL as the per-token reward,
and are trained for 150 steps with 1,024 prompts per step;
the scheduling methods start from $\mathbf{q}^{(0)}=(0.4, 0.4, 0.2)$ and update it every 10 steps, except Open-MOPD
and Vanilla DRO-MOPD, which keep this sampling mixture and reweight the per-domain rewards and losses, respectively.
Following \dthree{}-MOPD, we report the mean and standard deviation over three seeds under two checkpoint selection criteria.
\textit{Best-Overall Performance} (BO) takes the single checkpoint with the best mean domain score of each run,
and \textit{Best-per-Benchmark Performance} (BB) takes the best checkpoint on each benchmark.
As a scale-free summary, we also report the average per-benchmark rank (Avg. Rank; details in Appendix~\ref{app:protocol}).

\subsection{Overall Performance (RQ1)}
\label{sec:exp-rq1}

As shown in Table~\ref{tab:main}, we have the following threefold observations.

(1) \textbf{Training-free model merging can hardly fuse the capabilities of multiple teachers.}
ParamMerge-Avg ranks last on all six benchmarks, and ParamMerge-TA still falls behind all the adaptive
MOPD methods (Avg. Rank of 5.00 and 5.33 under BO and BB).
This is reasonable as the domain-specific RL updates of the teachers conflict and partly cancel out
when merged in the parameter space, whereas MOPD aligns the student with each teacher token by token
on its own responses.
Yet MOPD alone does not guarantee this advantage:
Vanilla MOPD with a fixed mixture even ranks below ParamMerge-TA (5.83 vs.\ 5.00 under BO),
whereas the adaptive methods (Open-MOPD, \dthree{}-MOPD, and \rdms{}) outperform Vanilla MOPD on all six benchmarks under both criteria.
This indicates that the key lies in allocating the domain budget.

(2) \textbf{Vanilla DRO-MOPD is unstable under noisy training signals.}
Although it up-weights the worst domain, Vanilla DRO-MOPD falls behind the progress-driven Open-MOPD
and \dthree{}-MOPD on four of six benchmarks (AIME24, AIME25, LCB v6, and IFEval),
with an Avg. Rank of 3.42 and 3.33 under BO and BB.
This is mainly because, besides the three instabilities of \S\ref{sec:method-dro}, reweighting the losses under a fixed sampling mixture only rescales the gradients and amplifies the KL noise of the up-weighted domain.
This indicates that robust scheduling should protect the domain regressing from its own best level rather than the one
with the largest loss.

(3) \textbf{Our \rdms{} consistently outperforms the existing methods.}
\rdms{} ranks first on five of six benchmarks with the lowest Avg. Rank of 1.50 under both BO and BB,
clearly ahead of \dthree{}-MOPD and Open-MOPD (2.58--2.67).
The gains are largest on math, i.e., 1.95 and 1.19 points over the best baseline on AIME24 and AIME25 under BO,
and also hold on code (e.g., 0.70 points on LCB v6) and IFEval.
The only exception is IFBench, whose 58 constraint types are designed to be out-of-distribution~\citep{DBLP:journals/corr/abs-2507-02833}
and are all absent from our IF training prompts, which may limit the adaptability of the distilled student to them.
Even so, all adaptive MOPD methods, including \rdms{}, exceed the IF teacher (22.50) on IFBench,
whereas on the in-distribution IFEval, \rdms{} achieves the best result.
\rdms{} also surpasses the domain teachers (RouteRL) on all six benchmarks,
with all seed standard deviations within 0.30, whereas those of \dthree{}-MOPD reach 0.81 and 0.74 on AIME24 and LCB v5 under BO.
This is mainly because the existing progress-driven methods act on progress scores estimated
from noisy KL curves (e.g., \dthree{}-MOPD normalizes its softmax by the maximum score).
Hence, noise-induced leader switches (Appendix~\ref{app:dyn}) and transient KL spikes can pull a large share of prompts toward one domain.
The three modules of \rdms{} bound exactly these effects.
This indicates that the bottleneck of adaptive scheduling is robustness to noisy signals rather than adaptivity itself.
\rdms{} outperforms every competing method on at least 15 of the 18 benchmark-seed pairs.

\begin{table}[tp]
  \centering
  \caption{Ablation study of \rdms{} (\%), with the relative change compared to \rdms{} in parentheses (variants are single runs with seed 0).}
  \label{tab:ablation}
  \footnotesize
  \setlength{\tabcolsep}{5.8pt}
  \renewcommand{\arraystretch}{0.95}
  \begin{adjustbox}{max width=\columnwidth}
  \begin{tabular}{c|c|c c c}
    \specialrule{.1em}{0em}{0em}
    Benchmark & \cellcolor{high}\textbf{\rdms{}} & \texttt{w/o} \texttt{NRC} & \texttt{w/o} \texttt{SAE} & \texttt{w/o} \texttt{WDR} \\
    \hline
    \multicolumn{5}{c}{\textit{Best-Overall (BO)}} \\
    \hline
    AIME24 & \cellcolor{high}26.72 & 25.47 {\scriptsize\textcolor{gray}{($\downarrow$4.68\%)}} & 26.04 {\scriptsize\textcolor{gray}{($\downarrow$2.54\%)}} & 26.20 {\scriptsize\textcolor{gray}{($\downarrow$1.95\%)}} \\
    AIME25 & \cellcolor{high}26.09 & 25.62 {\scriptsize\textcolor{gray}{($\downarrow$1.80\%)}} & 25.36 {\scriptsize\textcolor{gray}{($\downarrow$2.80\%)}} & 25.89 {\scriptsize\textcolor{gray}{($\downarrow$0.77\%)}} \\
    LCB v5 & \cellcolor{high}23.79 & 22.93 {\scriptsize\textcolor{gray}{($\downarrow$3.61\%)}} & 23.47 {\scriptsize\textcolor{gray}{($\downarrow$1.35\%)}} & 23.65 {\scriptsize\textcolor{gray}{($\downarrow$0.59\%)}} \\
    LCB v6 & \cellcolor{high}23.01 & 22.51 {\scriptsize\textcolor{gray}{($\downarrow$2.17\%)}} & 23.37 {\scriptsize\textcolor{gray}{($\uparrow$1.56\%)}} & 22.86 {\scriptsize\textcolor{gray}{($\downarrow$0.65\%)}} \\
    IFEval & \cellcolor{high}74.74 & 74.58 {\scriptsize\textcolor{gray}{($\downarrow$0.21\%)}} & 74.82 {\scriptsize\textcolor{gray}{($\uparrow$0.11\%)}} & 75.74 {\scriptsize\textcolor{gray}{($\uparrow$1.34\%)}} \\
    IFBench & \cellcolor{high}23.08 & 22.75 {\scriptsize\textcolor{gray}{($\downarrow$1.43\%)}} & 22.50 {\scriptsize\textcolor{gray}{($\downarrow$2.51\%)}} & 23.00 {\scriptsize\textcolor{gray}{($\downarrow$0.35\%)}} \\
    \hline
    \multicolumn{5}{c}{\textit{Best-per-Benchmark (BB)}} \\
    \hline
    AIME24 & \cellcolor{high}26.72 & 26.30 {\scriptsize\textcolor{gray}{($\downarrow$1.57\%)}} & 26.04 {\scriptsize\textcolor{gray}{($\downarrow$2.54\%)}} & 26.20 {\scriptsize\textcolor{gray}{($\downarrow$1.95\%)}} \\
    AIME25 & \cellcolor{high}26.09 & 25.62 {\scriptsize\textcolor{gray}{($\downarrow$1.80\%)}} & 25.36 {\scriptsize\textcolor{gray}{($\downarrow$2.80\%)}} & 25.89 {\scriptsize\textcolor{gray}{($\downarrow$0.77\%)}} \\
    LCB v5 & \cellcolor{high}23.95 & 23.59 {\scriptsize\textcolor{gray}{($\downarrow$1.50\%)}} & 23.47 {\scriptsize\textcolor{gray}{($\downarrow$2.00\%)}} & 23.65 {\scriptsize\textcolor{gray}{($\downarrow$1.25\%)}} \\
    LCB v6 & \cellcolor{high}23.20 & 23.20 {\scriptsize\textcolor{gray}{(0.00\%)}} & 23.37 {\scriptsize\textcolor{gray}{($\uparrow$0.73\%)}} & 22.86 {\scriptsize\textcolor{gray}{($\downarrow$1.47\%)}} \\
    IFEval & \cellcolor{high}74.88 & 74.58 {\scriptsize\textcolor{gray}{($\downarrow$0.40\%)}} & 74.82 {\scriptsize\textcolor{gray}{($\downarrow$0.08\%)}} & 75.74 {\scriptsize\textcolor{gray}{($\uparrow$1.15\%)}} \\
    IFBench & \cellcolor{high}23.17 & 22.75 {\scriptsize\textcolor{gray}{($\downarrow$1.81\%)}} & 23.42 {\scriptsize\textcolor{gray}{($\uparrow$1.08\%)}} & 23.08 {\scriptsize\textcolor{gray}{($\downarrow$0.39\%)}} \\
    \specialrule{.1em}{0em}{0em}
  \end{tabular}
  \end{adjustbox}
\end{table}

\subsection{Empirical Module Analyses (RQ2)}
\label{sec:exp-rq2}

\subsubsection{\textbf{Ablation Study}}
\label{sec:exp-ablation}
To verify each module, we build three variants of \rdms{},
each disabling exactly one module by its hyper-parameter (\S\ref{sec:method-alg})
and keeping all other hyper-parameters unchanged.
As shown in Table~\ref{tab:ablation}, disabling any module degrades \rdms{} on most benchmarks.
\begin{itemize}[leftmargin=*]
  \item \underline{\texttt{w/o} \texttt{NRC}} drops on all six benchmarks under BO,
    most notably by 4.68\% and 3.61\% on AIME24 and LCB v5,
    as the scheduler without the noise floor pushes the budget toward a domain that leads by chance, i.e., the worst-case misjudgment.
    This implies the criticalness of deciding \textit{when} to respond under noisy signals.
  \item \underline{\texttt{w/o} \texttt{SAE}} drops on four of six benchmarks under both criteria,
    most notably by 2.54\% and 2.80\% on AIME24 and AIME25.
    Without clipping, transient KL spikes of the code domain enter the statistics of the nominal allocation
    in proportion to their size, distorting the allocation and squeezing the other domains.
    This implies the criticalness of bounding the influence of any single anomaly.
  \item \underline{\texttt{w/o} \texttt{WDR}} drops on all benchmarks except IFEval under both criteria,
    e.g., by 1.95\% on AIME24 and 1.47\% on LCB v6 under BB,
    as the nominal allocation gives no extra protection to a domain regressing from its best level.
    This implies the criticalness of each domain's own best level as the reference.
\end{itemize}

\subsubsection{\textbf{Module Stability Analyses}}
\label{sec:exp-modules}

\begin{figure}[t]
  \centering
  \includegraphics[width=\linewidth]{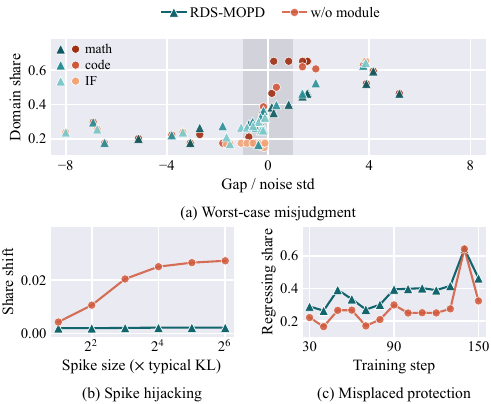}
  \caption{\rdms{} vs.\ its variants without the corresponding module (``w/o module''), replayed on the same
  recorded trajectory.
  (a) Share of each domain vs.\ its progress gap in noise standard deviations
  (shaded: within one std);
  (b) share shift remaining 40 steps after a KL spike (lower is better);
  (c) share of the most regressing domain (higher is better).}
  \label{fig:module}
\end{figure}

To understand why each module works, we examine whether it resolves its instability of Vanilla DRO-MOPD
(\S\ref{sec:method-dro}).
We replay \rdms{} and its variants without each module on the same recorded trajectory,
so that any difference in decisions is attributable to this module alone.
As shown in Figure~\ref{fig:module}, we have the following threefold observations.

(1) \textbf{\texttt{NRC} resolves the worst-case misjudgment caused by noise.}
As shown in Figure~\ref{fig:module}(a), without \texttt{NRC}, the shares swing between 0.15 and 0.65
even when the gap between domains is below one noise standard deviation, i.e., the worst-case misjudgment.
With \texttt{NRC}, they stay between 0.17 and 0.40 within the noise range, yet still rise to 0.57 on average
once the gap exceeds two noise standard deviations.
As a result, the share follows the gap much more closely (Spearman correlation of 0.83 vs.\ 0.55).
This indicates that \rdms{} responds only to real learning gaps, while still adjusting the mixture decisively.

(2) \textbf{\texttt{SAE} prevents spikes from hijacking the training budget.}
As shown in Figure~\ref{fig:module}(b), without \texttt{SAE}, the share shift caused by an injected one-step
KL spike grows with the spike size and lasts long after the spike disappears, i.e., the spike hijacking.
The shift is still 0.027 after 40 steps under a 64$\times$ spike.
With \texttt{SAE}, it stays at 0.002 regardless of the spike size,
and its maximum right after the spike is bounded at 0.033, compared with 0.172 without \texttt{SAE}.
This indicates that the influence of any single anomaly on \rdms{} is both bounded and short-lived.

(3) \textbf{\texttt{WDR} protects the truly regressing domains.}
As shown in Figure~\ref{fig:module}(c), without \texttt{WDR}, the domain regressing the most from its own
best level receives only 0.277 of the budget on average, even less than the 0.398 of the domains close to
their best levels, i.e., the misplaced protection.
With \texttt{WDR}, it receives more budget at 12 of the 13 updates from step 30, and its share rises to 0.382 on average.
This indicates that \rdms{} protects the truly regressing domains.

\subsection{Robustness under a Corrupted Teacher (RQ3)}
\label{sec:exp-rq3}

\begin{figure}[t]
  \centering
  \includegraphics[width=\linewidth]{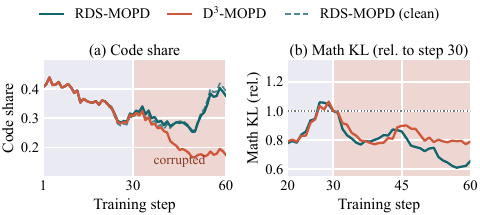}
  \caption{\rdms{} vs.\ \dthree{}-MOPD with the code teacher corrupted from step 30 (shaded).
  (a) Share of code prompts (dashed: \rdms{} without corruption);
  (b) math KL relative to its level at step 30 (lower is better).}
  \label{fig:stress}
\end{figure}

In practice, the signal of a domain teacher may become unreliable during training, e.g., due to a faulty scoring service.
To test whether \rdms{} stays reliable in such cases, we corrupt the code teacher from step 30 on.
Specifically, we add zero-mean noise (re-sampled at every call) to its token log-probabilities within the
student's top-16 candidates, which inflates its KL signal by about 2.5$\times$.
We then compare \rdms{} with \dthree{}-MOPD under the same seed and data stream for 60 steps,
which cover the first three scheduling updates after the corruption.
As shown in Figure~\ref{fig:stress}, we have the following twofold observations.

(1) \textbf{\rdms{} keeps a stable mixture under the corrupted teacher.}
As shown in Figure~\ref{fig:stress}(a), the mixture of \rdms{} is almost unaffected by the corruption.
Its code share deviates from that of its uncorrupted run by only 0.011 on average over steps 30--60,
and averages 0.325 over steps 41--60.
In contrast, \dthree{}-MOPD allocates only 0.179 to code and moves the released budget almost entirely to math
(+0.145 vs.\ $-$0.006 for \rdms{}).
This is mainly because the inflated KL raises the excess loss of code, so \texttt{WDR} keeps a bounded protection on it.
Meanwhile, \texttt{NRC} treats the remaining score differences within the noise level as no evidence for re-allocation,
so \rdms{} is not dragged by an unreliable teacher.

(2) \textbf{\rdms{} preserves the learning progress of all domains.}
\rdms{} allocates about 1.8$\times$ as many prompts to the corrupted code domain as \dthree{}-MOPD,
yet ends with a lower code KL (0.0164 vs.\ 0.0225 over steps 56--60).
Meanwhile, as shown in Figure~\ref{fig:stress}(b), its math KL decreases to 0.66$\times$ its level at step 30,
compared with 0.79$\times$ for \dthree{}-MOPD.
Thus, \rdms{} neither abandons the corrupted domain nor sacrifices clean ones.

\subsection{Efficiency Analysis (RQ4)}
\label{sec:exp-rq4}

\begin{figure}[t]
  \centering
  \includegraphics[width=0.88\linewidth]{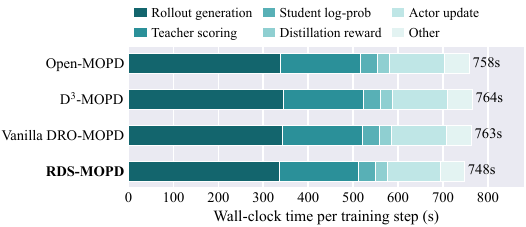}
  \caption{Wall-clock time per training step of MOPD methods.}
  \label{fig:overhead}
\end{figure}

We measure the per-step wall-clock time of the MOPD methods on the same hardware (16 A800 GPUs; stages in
Figure~\ref{fig:overhead}) and the time of the scheduling decision, with the following twofold observations.

(1) \textbf{The scheduling decision of \rdms{} introduces negligible overhead.}
The \rdms{} scheduler takes only 0.039 milliseconds per step on average,
more than seven orders of magnitude below a training step of about 750 seconds.
This is mainly because all three modules operate on a few per-domain scalars, with no extra forward pass
or GPU memory.
Thus, the robustness of \rdms{} comes from its decision rule, not extra computation.

(2) \textbf{The end-to-end training cost of \rdms{} is on par with or even lower than that of existing methods.}
Over steps 2--84, \rdms{} takes 747.5 seconds per step on average,
vs.\ 758.2, 764.5, and 762.6 seconds for Open-MOPD, \dthree{}-MOPD, and Vanilla DRO-MOPD.
Over the full 150 steps, it is 5.6\% faster per step than \dthree{}-MOPD,
as rollout takes about 45\% of each step and the student of \rdms{} produces shorter responses
(7{,}130 vs.\ 8{,}127 tokens on average).
Thus, \rdms{} distills better at no extra cost.

%% file: sections/7_related.tex
\section{Related Work}
\label{sec:related}

\subsection{Multi-Teacher and On-Policy Distillation}

Knowledge distillation fits a student to the teacher's output distribution~\citep{DBLP:journals/corr/HintonVD15},
and early multi-teacher methods fuse the soft labels or intermediate representations of multiple
teachers~\citep{DBLP:conf/kdd/YouX0T17}, or select the teachers for each sample by reinforcement
learning~\citep{DBLP:conf/aaai/YuanSPLGFJ21}.
For LLMs, sequence-level distillation~\citep{DBLP:conf/emnlp/KimR16} suffers from
the distribution shift between training and inference,
which on-policy distillation (OPD) alleviates by supervising the student on its own sampled
responses~\citep{DBLP:conf/iclr/AgarwalVZSGGB24,DBLP:conf/iclr/GuDWH24}, e.g., with the reverse KL divergence
to avoid overestimating the low-probability regions of the teacher~\citep{DBLP:conf/iclr/GuDWH24},
and has become a standard post-training
stage~\citep{TML:LuOnPolicyDistillation25,DBLP:journals/corr/abs-2505-09388},
as its per-token supervision is denser than the sequence-level rewards of RL.
Extending it to multiple domain-expert teachers leads to MOPD~\citep{DBLP:journals/corr/abs-2606-30406},
which merges the capabilities of domain-specific RL teachers into one student.
Subsequent work improves MOPD by balancing the capabilities of different
teachers via token-share balancing and KL-gap following~\citep{DBLP:journals/corr/abs-2608-19098},
weighting or gating teacher supervision by its
reliability~\citep{DBLP:journals/corr/abs-2609-23697,DBLP:journals/corr/abs-2609-02998},
resolving the conflicts among
teachers~\citep{DBLP:journals/corr/abs-2610-13521,DBLP:journals/corr/abs-2605-27115},
and distilling at the representation level~\citep{DBLP:journals/corr/abs-2610-02381}.
Most of them keep the domain mixture fixed and refine how the teacher supervision is weighted or
combined.
Closer to our work, D$^3$-MOPD~\citep{DBLP:journals/corr/abs-2608-24987} schedules the domain mixture by the learning
progress of each domain, i.e., its remaining KL gap times its descent velocity,
MT-GRPO~\citep{DBLP:conf/icml/RameshJZYWBLB26} reweights tasks online toward the worst-performing task in
multi-task RL, and concurrent work~\citep{DBLP:journals/corr/abs-2610-26525} sets model-merging
coefficients before training by minimizing the worst per-domain residual.
Different from these works, \rdms{} introduces DRO into the online domain-mixture scheduling of
MOPD and makes it robust to noise.

\subsection{Distributionally Robust Optimization}

DRO optimizes the worst-case loss within an uncertainty set around the nominal
distribution~\citep{DBLP:journals/mansci/Ben-TalHWMR13,DBLP:journals/aos/DuchiN21},
where the uncertainty set can be defined by, e.g.,
the KL divergence, whose worst case admits a closed-form exponential
tilting~\citep{DBLP:journals/mor/Lam16},
the Wasserstein distance~\citep{DBLP:journals/mp/EsfahaniK18},
or a CVaR-type tail risk~\citep{RockafellarU00}.
Group DRO~\citep{DBLP:conf/iclr/SagawaKHL20} instantiates this idea over predefined groups by up-weighting
high-loss groups online, and has been widely adopted,
e.g., in topic-robust language modeling~\citep{DBLP:conf/emnlp/OrenSHL19},
data mixture optimization for pre-training~\citep{DBLP:conf/nips/XiePDLLLLLY23},
group-robust preference alignment~\citep{DBLP:conf/nips/RameshHCMSAB24},
and worst-group knowledge
distillation~\citep{DBLP:journals/tmlr/KenfackAK25,DBLP:conf/uai/WangNZHLM23,DBLP:conf/miccai/VilourasLSOT23}.
Among them, DoReMi~\citep{DBLP:conf/nips/XiePDLLLLLY23} runs Group DRO on a small proxy model to determine the
pre-training domain weights offline.
However, DRO is known to be sensitive to outliers,
since the worst case tends to be dominated by them, and DORO~\citep{DBLP:conf/icml/ZhaiDKR21} thus discards a
fraction of the highest-loss samples.
This motivates combining DRO with robust statistics against heavy-tailed noise, e.g., the Huber estimator~\citep{Huber1964}
and median-of-means~\citep{DBLP:journals/focm/LugosiM19}.
Following this line, \rdms{} fixes three concrete instabilities of DRO under the noisy signals of MOPD
within a single DRO structure.

%% file: sections/8_conclusion.tex
\section{Conclusion and Future Work}
\label{sec:conclusion}
In this paper, we propose \rdms{}, a robust domain-mixture scheduling framework for multi-teacher on-policy distillation.
It resolves the instabilities of Vanilla DRO-MOPD under noisy signals with three modules, backed by theoretical guarantees.
Extensive experiments on mathematical reasoning, code generation, and instruction-following benchmarks demonstrate
the effectiveness and robustness of \rdms{} with negligible computational overhead.
As for future work, we plan to extend the robust allocation from the domain level to finer granularities,
e.g., the prompt level.

%% file: sections/9_appendix.tex
\appendix

\section{Notations}
\label{app:notation}

\begin{table}[h]
  \centering
  \caption{Notations and meanings.}
  \label{tab:notation}
  \renewcommand{\arraystretch}{1.0}
  \begin{adjustbox}{max width=\linewidth}
  \begin{tabular}{c|l}
    \specialrule{.1em}{.05em}{.05em}
    \textbf{Notation} & \textbf{Meaning} \\
    \hline
    $\Delta^{K},\ \mathbf{u},\ \mathbf{1},\ \odot$ & \begin{tabular}[c]{@{}l@{}}probability simplex; uniform allocation $\mathbf{1}/K$;\\all-ones vector; element-wise product\end{tabular} \\
    \hline
    $\mathcal{D},\ d,\ K$ & domain set, a domain, number of domains ($K=|\mathcal{D}|$) \\
    \hline
    $\pi_\theta,\ \pi_{\phi_d},\ \mathcal{P}_d$ & student; RL teacher routed to domain $d$; prompt pool of $d$ \\
    \hline
    $x,\ y,\ |y|,\ j$ & prompt; on-policy student response; its length; token position \\
    \hline
    $D_j,\ k,\ \mathcal{L}_d(\theta)$ & \begin{tabular}[c]{@{}l@{}}top-$k$ reverse KL at position $j$ (reward $-D_j$);\\top-$k$ size; distillation loss of $d$\end{tabular} \\
    \hline
    $s,\ t,\ \Delta,\ B,\ S$ & \begin{tabular}[c]{@{}l@{}}training step; scheduling update;\\update interval; prompts per batch; total steps\end{tabular} \\
    \hline
    $m_d^{(s)},\ \tilde m_d^{(s)},\ \mathcal{B}_d$ & \begin{tabular}[c]{@{}l@{}}per-domain KL observation of $d$ at step $s$; its clipped value\\in \texttt{SAE}; domain-$d$ responses in the batch\end{tabular} \\
    \hline
    $\bar m_d^{(s)},\ \bar m_d^{\star},\ \xi_d^{(s)}$ & true KL level of $d$ at step $s$; best true level so far; log-noise \\
    \hline
    $\mathbf{q}^{(t)},\ \mathbf{p}^{(t)}$ & \begin{tabular}[c]{@{}l@{}}domain mixture at update $t$;\\nominal allocation (progress-driven in \rdms{})\end{tabular} \\
    \hline
    $\hat\ell_d^{(s)}$ & loss estimate in Eqn.~\eqref{eq:dro} ($m_d^{(s)}/I_d$ in Vanilla DRO-MOPD) \\
    \hline
    $I_d,\ \ell_d,\ \ell_d^{\star}$ & initial KL level; current EMA level; best (lowest) level so far \\
    \hline
    $g_d,\ v_d,\ r_d,\ r_{\max}$ & \begin{tabular}[c]{@{}l@{}}remaining gap; descent velocity;\\progress score ($r_d=g_dv_d$); largest score\end{tabular} \\
    \hline
    $W,\ n_0,\ K_p,\ J$ & EMA window; points for $I_d$; max./used rolling deltas \\
    \hline
    $\ell_d^{[j]},\ \boldsymbol{\omega},\ \varrho_{d,j}$ & EMA of window $j$; EMA weights; ratio $\ell_d^{[j]}/\ell_d^{[j+1]}$ \\
    \hline
    $T,\ \tau^{(t)},\ f,\ \alpha,\ z_d$ & base/effective temperature; share floor; jitter range/value \\
    \hline
    $\sigma_d,\ \hat\sigma^{(t)},\ N_\sigma,\ \epsilon$ & \begin{tabular}[c]{@{}l@{}}log-noise scale; SE of the leading score;\\noise window; noise-floor multiplier\end{tabular} \\
    \hline
    $\mu_d^{(s)},\ H,\ c$ & running median; median window; Huber clipping threshold \\
    \hline
    $h_d,\ \kappa,\ \eta,\ \lambda_{\mathrm{lo}}$ & excess loss; excess cap; DRO (tilt) strength; coverage floor \\
    \hline
    $\mathcal{U},\ \lambda_d,\ \nu$ & feasible set; per-domain floor; normalizing constant \\
    \hline
    $\zeta,\ s_0,\ M$ & score bound in SEs; step of an anomaly or jump; anomaly size \\
    \hline
    $\rho,\ \gamma$ & KL radius; dual variable (proof of Theorem~\ref{thm:tilt}) \\
    \specialrule{.1em}{.05em}{.05em}
  \end{tabular}
  \end{adjustbox}
\end{table}

\section{Benchmark and Training Data Details}
\label{app:benchmarks}
Table~\ref{tab:bench} lists the six benchmarks: AIME24/25~\citep{MAA:AIME} contain the 2024/2025 AIME problems with integer answers judged by exact match.
LiveCodeBench v5/v6~\citep{DBLP:conf/iclr/JainHGLYZWSSS25} collect competitive programming problems by release
window and judge each solution by hidden test cases, which alleviates contamination.
IFEval~\citep{DBLP:journals/corr/abs-2311-07911} contains programmatically verifiable constraints (e.g., length,
format, and keywords), and IFBench~\citep{DBLP:journals/corr/abs-2507-02833} adds novel out-of-distribution
constraints; a response is correct only if it satisfies all constraints under the strict prompt-level criterion.
We score each IF sample against its own response, unlike the released script of
Open-MOPD~\citep{DBLP:journals/corr/abs-2608-19098}, which gives all samples of a prompt one verdict.
All metrics are pass@1 estimated by avg@\#Samp., with thinking mode, top-$p=0.95$, and at most 30k generated tokens,
and each run is summarized over its checkpoints by BO and BB (Appendix~\ref{app:protocol}).

We build our testbed on Open-MOPD~\citep{DBLP:journals/corr/abs-2608-19098}, to our knowledge the only recent MOPD work
that releases its complete training resources (SFT student, domain teachers, and prompts), so all methods share the same student, teachers, and data.
The student is SmolLM3-3B~\citep{HF:SmolLM3} initialized from the released MixSFT checkpoint (Student-SFT), and
the teachers RL-Math, RL-Code, and RL-IF are obtained by domain-specific RL on top of Student-SFT.
The prompt pool (Table~\ref{tab:traindata}) draws math from DAPO~\citep{DBLP:journals/corr/abs-2503-14476},
code from PrimeIntellect~\citep{PrimeIntellect:SYNTHETIC1} and TACO~\citep{DBLP:journals/corr/abs-2312-14852},
and IF from Nemotron-IF-RL~\citep{DBLP:journals/corr/abs-2608-19098}.
All trained methods share this pool and differ mainly in the domain mixture $\mathbf{q}^{(t)}$ (except Vanilla DRO-MOPD, Appendix~\ref{app:impl}).

\begin{table}[h]
  \centering
  \caption{Benchmarks and training data per domain (\#Prob.: problems; \#Samp.: samples per problem, scored by
  avg@\#Samp.; Temp.: sampling temperature).}
  \label{tab:bench}
  \label{tab:traindata}
  \footnotesize
  \setlength{\tabcolsep}{1.9pt}
  \begin{adjustbox}{max width=\columnwidth}
  \begin{tabular}{c|c c c c|c c c}
    \specialrule{.1em}{0em}{0em}
    \multirow{2}{*}{Domain} & \multicolumn{4}{c|}{Evaluation} & \multicolumn{3}{c}{Training} \\
      & Benchmark & \#Prob. & \#Samp. & Temp. & Source & \#Prompts & Teacher \\
    \hline
    \multirow{2}{*}{Math} & AIME24 & 30 & 64 & 0.6 & \multirow{2}{*}{DAPO} & \multirow{2}{*}{17,917} & \multirow{2}{*}{RL-Math} \\
                          & AIME25 & 30 & 64 & 0.6 & & & \\
    \hline
    \multirow{2}{*}{Code} & LCB v5 & 167 & 10 & 1.0 & \multirow{2}{*}{\shortstack{PrimeIntellect\\+ TACO}} & \multirow{2}{*}{23,667} & \multirow{2}{*}{RL-Code} \\
                          & LCB v6 & 175 & 10 & 1.0 & & & \\
    \hline
    \multirow{2}{*}{IF}   & IFEval  & 541 & 4 & 1.0 & \multirow{2}{*}{\shortstack{Nemotron-\\IF-RL}} & \multirow{2}{*}{45,347} & \multirow{2}{*}{RL-IF} \\
                          & IFBench & 300 & 4 & 1.0 & & & \\
    \specialrule{.1em}{0em}{0em}
  \end{tabular}
  \end{adjustbox}
\end{table}

\section{Baseline Details}
\label{app:baselines}
\textbf{Vanilla MOPD}~\citep{DBLP:journals/corr/abs-2606-30406,DBLP:journals/corr/abs-2608-24987} fixes the
mixture to each domain's share of the prompt pool.
\textbf{Open-MOPD}~\citep{DBLP:journals/corr/abs-2608-19098} adds token-share balancing, KL-gap following, and a reward refresh within every inner update.
\textbf{Vanilla DRO-MOPD}~\citep{DBLP:conf/iclr/SagawaKHL20} normalizes each per-domain loss by its initial level;
its strength $\eta$ is set as in Appendix~\ref{app:impl}.
\textbf{\dthree{}-MOPD}~\citep{DBLP:journals/corr/abs-2608-24987} samples each domain by the softmax of its remaining KL gap times its descent velocity, with a share floor.
\textbf{ParamMerge-Avg}~\citep{DBLP:conf/icml/WortsmanIGRLMNF22} averages the parameters of the three teachers.
\textbf{ParamMerge-TA}~\citep{DBLP:conf/iclr/IlharcoRWSHF23} adds the scaled sum of the teachers' task vectors
(differences from the SFT student) to the student; we report the better of scaling factors 0.5 and 1.0.
The reference rows of Table~\ref{tab:main} are excluded from the ranking: \textbf{Student-SFT} is the starting
point of all trained methods, and \textbf{RouteRL}~\citep{DBLP:journals/corr/abs-2608-19098} answers each prompt
with its domain teacher (RL-Math for AIME24/25, RL-Code for LCB v5/v6, and RL-IF for IFEval/IFBench), which needs an oracle router and all teachers.

\section{Implementation Details}
\label{app:impl}
\textbf{Training.}
All methods use the dense top-16 reverse KL of the routed teacher as the per-token reward, $B=1{,}024$ prompts
per step, a mini-batch size of 256, 150 steps, a learning rate of $1.5\times10^{-6}$, and at most 16k
response tokens.
Trained methods start from $\mathbf{q}^{(0)}=(0.4, 0.4, 0.2)$ (math : code : IF), except Vanilla MOPD
(pool-proportional), and the scheduling methods update the mixture every $\Delta=10$ steps; Open-MOPD
and Vanilla DRO-MOPD reweight the per-domain rewards and losses instead (\S\ref{sec:method-dro-form}).
Each batch jitters the mixture as $\tilde q_d\propto q_d(1+z_d)$ with
$z_d\sim\mathrm{Unif}[-\alpha,\alpha]$ and draws $B\tilde q_d$ prompts by the largest-remainder rule.
All MOPD methods, including the ablation variants, share one training code and sampler and differ mainly in the mixture rule.

\textbf{Hyper-parameters.}
\rdms{} uses $\epsilon=4$, $N_\sigma=40$ (\texttt{NRC}), $c=0.5$, $H=5$ (\texttt{SAE}), $\eta=1$, $\kappa=2$,
$\lambda_{\mathrm{lo}}=0.2$ (\texttt{WDR}), $W=10$, $n_0=5$, $K_p=3$, $T=0.5$,
$f=0.10$, and $\alpha=0.30$ (sensitivity in Appendix~\ref{app:sens}).
Baselines use their official settings; Vanilla DRO-MOPD updates the domain weights by exponentiated gradient
on the per-domain loss normalized by its initial level, with strength $\eta=0.03$.

\textbf{Scheduler Formulas.}
\label{app:sched}
The progress score $r_d=g_d v_d$ follows \dthree{}-MOPD~\citep{DBLP:journals/corr/abs-2608-24987}: $g_d=\ell_d/I_d$,
and $v_d$ averages the relative decrease of $\ell_d^{[j]}$, the EMA over the $j$-th most recent window of $W$ clipped
observations ($\ell_d=\ell_d^{[0]}$), for $j<J\le K_p$, clipped at $0$.
\texttt{NRC} estimates the log-noise scale over the last $N_\sigma$ observations as
$\sigma_d=\frac{1.4826}{\sqrt{2}}\,\mathrm{median}_{s'}|\log\tilde m_d^{(s')}-\log\tilde m_d^{(s'-1)}|$,
and the delta method gives
\begin{equation}
  \begin{aligned}
  \mathrm{SE}(v_d)&=\frac{\sqrt{2}\,\sigma_d\|\boldsymbol{\omega}\|_2}{J}\Big(\sum_{j=0}^{J-1}\varrho_{d,j}^2\Big)^{1/2},\\
  \mathrm{SE}(r_d)&=\Big(\big(v_d\,\sigma_d\|\boldsymbol{\omega}\|_2\,g_d\big)^2+\big(g_d\,\mathrm{SE}(v_d)\big)^2\Big)^{1/2},
  \end{aligned}
  \label{eq:nf-se}
\end{equation}
where $\boldsymbol{\omega}$ are the EMA weights over one window and $\varrho_{d,j}=\ell_d^{[j]}/\ell_d^{[j+1]}$;
$\hat\sigma=\mathrm{SE}(r_{d_{\max}})$ with $d_{\max}=\arg\max_d r_d$.
During warm-up ($s\le H$), the median of \texttt{SAE} uses the available observations.
\texttt{WDR} tilts and normalizes $\mathbf{p}$, then iteratively clamps the coordinates below their floors and
rescales the others, which returns Eqn.~\eqref{eq:tilt} exactly within $K$ rounds (Appendix~\ref{app:proof-tilt}).

\begin{algorithm}[t]
  \caption{Training procedure of \rdms{}}
  \label{alg:rds}
  \small
  \KwIn{Student $\pi_\theta$; teachers $\{\pi_{\phi_d}\}$; pools $\{\mathcal{P}_d\}$; batch size $B$; steps $S$;
  interval $\Delta$; hyper-parameters (Appendix~\ref{app:impl}).}
  \KwOut{Distilled student $\pi_\theta$.}
  $\mathbf{q}\leftarrow\mathbf{q}^{(0)}$\;
  \For{$s=1,\dots,S$}{
    Sample $B$ prompts from $\{\mathcal{P}_d\}$ by the jittered $\mathbf{q}$\;
    Roll out, score with the routed teachers, and update $\pi_\theta$ with the reward $-D_j$ (\S\ref{sec:pre-mopd})\;
    \ForEach{$d\in\mathcal{D}$}{
      Read $m_d^{(s)}$ by Eqn.~\eqref{eq:kl-signal} and clip it into $\tilde m_d^{(s)}$ by Eqn.~\eqref{eq:hr} \tcp*{\texttt{SAE}}
    }
    \If{$s$ is a scheduling step (every $\Delta$ steps after $2W$ observations per domain)}{
      Get $\ell_d$, $v_d$, $r_d$, and $\hat\sigma$ on $\{\tilde m_d\}$ (Appendix~\ref{app:sched})\;
      Get $\tau$ by Eqn.~\eqref{eq:nf-temp} and $\mathbf{p}$ by Eqn.~\eqref{eq:rds-nominal} \tcp*{\texttt{NRC}}
      Update $\ell_d^{\star}$ and get $h_d$ by Eqn.~\eqref{eq:excess}\;
      $\mathbf{q}\leftarrow$ the solution of Eqn.~\eqref{eq:tilt} \tcp*{\texttt{WDR}}
    }
  }
  \Return{$\pi_\theta$}\;
\end{algorithm}

\textbf{Evaluation Setup.}
For each of the three seeds, we evaluate the checkpoints at steps 90, 100, 120, and 150, summarize them by the
BO and BB criteria (Appendix~\ref{app:protocol}), and apply the same rule to all methods.
The main runs use two nodes of eight NVIDIA A800-80GB GPUs, the ablation variants one such node, and the stress
test four GPUs, with Python 3.12, PyTorch 2.8.0 (CUDA 12.8), verl 0.7.0, vLLM 0.11.0, and Transformers 4.57.

\section{Evaluation Details}
\label{app:protocol}
Each benchmark is scored at every evaluated checkpoint (steps 90, 100, 120, and 150), and each run
is summarized by two criteria.
\textbf{Best-Overall (BO)} selects the single checkpoint with the best mean of the three domain scores (each
averaging its two benchmarks) and reports all six benchmarks there; it reflects one deployable model.
\textbf{Best-per-Benchmark (BB)} takes the best score of each benchmark over all checkpoints separately.
We report the mean and standard deviation over three seeds; training-free methods and reference rows have identical
results under both criteria.
As a scale-free summary, Avg. Rank ($\downarrow$) ranks all ranked methods on each
benchmark by their 3-seed means (1 is best, ties in the unrounded means averaged) and averages over the six benchmarks, per criterion.
For significance, as three seeds are too few, we run the one-sided Wilcoxon signed-rank test
between \rdms{} and each competitor over all 18 benchmark-seed pairs (six benchmarks $\times$ three seeds), per
criterion, treating pairs as independent.
\rdms{} wins on at least 15 out of the 18 pairs against every competitor.

\section{Mixture Dynamics}
\label{app:dyn}

\begin{figure}[t]
  \centering
  \includegraphics[width=\linewidth]{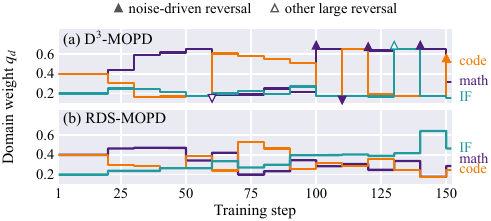}
  \caption{Domain shares during training (seed 0).
  Filled/open triangles: noise-driven/other large reversals of \dthree{}-MOPD, marked at the
  most-changed domain; \rdms{} has none.}
  \label{fig:dyn}
\end{figure}

We compare the real training runs of \rdms{} and \dthree{}-MOPD (same seed and hardware).
An update is a \textit{large share reversal} if any domain share changes by more than 0.3, and the
\textit{mean share change} of a run averages $\frac{1}{2}\|\mathbf{q}^{(t)}-\mathbf{q}^{(t-1)}\|_1$.
A large reversal is \textit{noise-driven} if the same scheduler reproduces it (same update, domain, and direction)
in fewer than half of 100 bootstrap replicas of the KL trajectory, resampling residuals around a 21-step running
median.

(1) \textbf{\dthree{}-MOPD oscillates the mixture with noise.}
It makes 7 large share reversals in 14 updates, 5 of which are noise-driven (reappearing in only 13--28\% of
the replicas), e.g., from step 100 to 140 the math share alternates between 0.175 and 0.651 (Figure~\ref{fig:dyn}(a)) while the math KL
stays within 0.002--0.006.
This is mainly because its temperature shrinks with the progress scores (from 0.086 at step 20 to about 0.006 after step
90), so batch-level KL fluctuations alone move most of the budget, i.e., the worst-case misjudgment of
\S\ref{sec:method-dro}.

(2) \textbf{\rdms{} keeps the mixture responsive yet stable.}
It makes no large share reversal and cuts the mean share change from 0.258 to 0.118 (54\%) and the maximum from 0.478
to 0.285 (Figure~\ref{fig:dyn}), while still following real gaps, e.g., the IF share rises to 0.643 at step 140 as only IF still descends.
This is mainly because \texttt{NRC} calibrates the radius to the estimated noise level.

\section{Hyper-parameter Sensitivity}
\label{app:sens}

\begin{figure}[t]
  \centering
  \includegraphics[width=\linewidth]{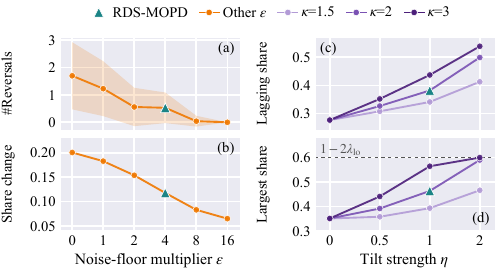}
  \caption{Sensitivity to $\epsilon$ of \texttt{NRC} (a, b) and $\eta,\kappa$ of \texttt{WDR} (c, d)
  ($\blacktriangle$: default $\epsilon=4$, $\eta=1$, $\kappa=2$): (a) noise-driven reversals per bootstrap replica
  (mean $\pm$ std); (b) mean share change per update; (c) lagging-domain share; (d) max final share
  (dashed: $1-2\lambda_{\mathrm{lo}}$, i.e., two domains at $\lambda_{\mathrm{lo}}$).}
  \label{fig:sens}
\end{figure}

We vary $\epsilon\in\{0, 1, 2, 4, 8, 16\}$ (\texttt{NRC}) and
$(\eta,\kappa)\in\{0, 0.5, 1, 2\} \times \{1.5, 2, 3\}$ (\texttt{WDR}) with the others at their defaults
(Appendix~\ref{app:impl}); $\epsilon=0$ and $\eta=0$ are \texttt{w/o} \texttt{NRC} and \texttt{w/o} \texttt{WDR}.
We replay the scheduler on the recorded KL trajectory of \rdms{} (150 steps, 14 updates; Figure~\ref{fig:sens}) to isolate the
scheduling decisions, with reversals as in Appendix~\ref{app:dyn}.
The \textit{lagging domain} at an update is the one whose KL regresses most from its own best level.

(1) \textbf{A moderate noise floor reduces noise-driven reversals without freezing the mixture.}
Raising $\epsilon$ from 0 to 4 cuts noise-driven reversals on the recorded trajectory from 3 to 0 and the mean
share change from 0.200 to 0.118 (41\%); $\epsilon=4$ is the smallest value removing all large reversals (5, 4,
and 3 for $\epsilon=0$, 1, and 2) while keeping a largest single-update change of 0.285, whereas $\epsilon=8$ and
16 (0.04 and 0.00 noise-driven reversals) suppress it to 0.158 and 0.161, making the scheduler sluggish.

(2) \textbf{The tilt strength $\eta$ controls how much budget goes to the lagging domain.}
Its share grows monotonically with $\eta$ under every $\kappa$, e.g., 0.277, 0.327, 0.382, and 0.499 for
$\eta=0$, 0.5, 1, and 2 under $\kappa=2$, so the default $\eta=1$ adds 0.105 at no extra reversals.

(3) \textbf{The excess cap $\kappa$ prevents over-concentration.}
Under $\eta=1$, the largest final share grows from 0.393 to 0.464 and 0.565 for $\kappa=1.5$, 2, and 3, and
$\eta=2$, $\kappa=3$ ends at (0.200, 0.200, 0.600) with two domains on the floor $\lambda_{\mathrm{lo}}=0.2$
and 1 noise-driven reversal, whereas the default $\kappa=2$ ends at (0.288, 0.249, 0.464).

%% file: sections/9b_proofs.tex
\section{Proofs}
\label{app:proofs}

\subsection{Proof of Proposition~\ref{prop:nf}}
\label{app:proof-nf}
Proposition~\ref{prop:nf} is part (i); the full statement adds
(ii) for a given $\hat\sigma>0$ and any $\mathbf{r},\mathbf{r}'\ge\mathbf{0}$,
$\|\mathbf{p}(\mathbf{r})-\mathbf{p}(\mathbf{r}')\|_1\le\frac{2(1-Kf)}{T\,\epsilon\,\hat\sigma}\|\mathbf{r}-\mathbf{r}'\|_\infty$;
and (iii) with $\epsilon=0$, Eqn.~\eqref{eq:nf-temp} reduces to the max-normalized temperature.
Under the max-normalized temperature, such scores can instead change a share by
$(1-Kf)\frac{e^{1/T}-1}{e^{1/T}+K-1}$ for any $\zeta>0$.

\begin{lemma}[$\ell_1$ Sensitivity of Softmax]
\label{lem:softmax}
For any $\mathbf{x},\mathbf{x}'\in\mathbb{R}^{K}$,
$\|\mathrm{softmax}(\mathbf{x})-\mathrm{softmax}(\mathbf{x}')\|_1\le\frac{1}{2}\,\mathrm{osc}(\mathbf{x}-\mathbf{x}')$, where $\mathrm{osc}(\mathbf{y})=\max_d y_d-\min_d y_d$.
Moreover, any $\boldsymbol{\Delta}\in\mathbb{R}^{K}$ with $\sum_d\Delta_d=0$ satisfies
$\max_d|\Delta_d|\le\frac{1}{2}\|\boldsymbol{\Delta}\|_1$.
\end{lemma}
\begin{proof}
Let $\boldsymbol{\delta}=\mathbf{x}'-\mathbf{x}$ and $\boldsymbol{\pi}(\beta)=\mathrm{softmax}(\mathbf{x}+\beta\boldsymbol{\delta})$
for $\beta\in[0,1]$; then $\frac{d}{d\beta}\pi_d(\beta)=\pi_d(\beta)\big(\delta_d-\langle\boldsymbol{\pi}(\beta),\boldsymbol{\delta}\rangle\big)$,
so $\|\frac{d}{d\beta}\boldsymbol{\pi}(\beta)\|_1=\mathbb{E}_{d\sim\boldsymbol{\pi}(\beta)}|\delta_d-\bar\delta|$
with $\bar\delta=\langle\boldsymbol{\pi}(\beta),\boldsymbol{\delta}\rangle$, the mean absolute deviation of
$X=\delta_d$ supported on $[\delta_{\min},\delta_{\max}]$ with $\delta_{\max}-\delta_{\min}=\mathrm{osc}(\boldsymbol{\delta})$ ($\delta_{\max}=\delta_{\min}$ is trivial).
By convexity, $(z-\bar\delta)^+\le(\delta_{\max}-\bar\delta)(z-\delta_{\min})/(\delta_{\max}-\delta_{\min})$ on $[\delta_{\min},\delta_{\max}]$, so
$\mathbb{E}|X-\bar\delta|=2\,\mathbb{E}(X-\bar\delta)^+\le 2(\delta_{\max}-\bar\delta)(\bar\delta-\delta_{\min})/(\delta_{\max}-\delta_{\min})\le(\delta_{\max}-\delta_{\min})/2$.
Integrating over $\beta$ gives the first claim; for the second, the positive and negative entries of $\boldsymbol{\Delta}$ each sum to
$\frac{1}{2}\|\boldsymbol{\Delta}\|_1$ in absolute value.
\end{proof}

\textbf{(i)}
Let $b^{(t')}=\epsilon\hat\sigma^{(t')}>0$ for $t'\in\{t-1,t\}$.
Since $r_{\max}^{(t')}\le\zeta\hat\sigma^{(t')}\le b^{(t')}$, Eqn.~\eqref{eq:nf-temp} gives
$\tau^{(t')}=T\,b^{(t')}$, so the softmax input $\mathbf{x}^{(t')}=\mathbf{r}^{(t')}/(T\epsilon\hat\sigma^{(t')})$
lies in $[0,\zeta/(\epsilon T)]^{K}$ and $\mathrm{osc}(\mathbf{x}^{(t)}-\mathbf{x}^{(t-1)})\le 2\zeta/(\epsilon T)$.
By Lemma~\ref{lem:softmax}, $\|\mathbf{p}^{(t)}-\mathbf{p}^{(t-1)}\|_1\le(1-Kf)\zeta/(\epsilon T)$,
so (both allocations sum to one) $|p_d^{(t)}-p_d^{(t-1)}|\le(1-Kf)\zeta/(2\epsilon T)$, i.e.,
$0.7\zeta/4=0.175\zeta$ for $K=3$, $f=0.1$, $T=0.5$, and $\epsilon=4$.
For the max-normalized response, $\mathbf{r}^{(t-1)}=(\delta,0,\dots,0)$ and
$\mathbf{r}^{(t)}=(0,\delta,0,\dots,0)$ with $0<\delta\le\zeta\hat\sigma$ normalize to $\mathbf{e}_1$ and
$\mathbf{e}_2$ for any $\delta$, so the first share drops from $f+(1-Kf)\frac{e^{1/T}}{e^{1/T}+K-1}=0.651$ to
$f+(1-Kf)\frac{1}{e^{1/T}+K-1}=0.175$ by default, i.e., by $(1-Kf)\frac{e^{1/T}-1}{e^{1/T}+K-1}=0.476$.

\textbf{(ii)}
Let $b=\epsilon\hat\sigma>0$, $n(\mathbf{r})=\max(\|\mathbf{r}\|_\infty,b)\ge b$, and
$\boldsymbol{\chi}(\mathbf{r})=\mathbf{r}/n(\mathbf{r})$, so that the softmax input is $\boldsymbol{\chi}(\mathbf{r})/T$
(note $\|\mathbf{r}\|_\infty=r_{\max}$ for $\mathbf{r}\ge\mathbf{0}$).
Then $\boldsymbol{\chi}(\mathbf{r})-\boldsymbol{\chi}(\mathbf{r}')=(\mathbf{r}-\mathbf{r}')/n(\mathbf{r})
+\mathbf{r}'\big(n(\mathbf{r}')-n(\mathbf{r})\big)/\big(n(\mathbf{r})n(\mathbf{r}')\big)$.
Both terms are bounded by $\|\mathbf{r}-\mathbf{r}'\|_\infty/b$ in $\ell_\infty$, the second since
$\|\mathbf{r}'\|_\infty/n(\mathbf{r}')\le 1$ and $n(\cdot)$ is $1$-Lipschitz in $\ell_\infty$.
Hence the oscillation of the softmax input difference is at most $4\|\mathbf{r}-\mathbf{r}'\|_\infty/(bT)$, and
Lemma~\ref{lem:softmax} with the factor $(1-Kf)$ gives the claim.
For the max-normalized response, replacing $b$ by $\min(r_{\max},r'_{\max})$ gives a constant that diverges as
the maximum score tends to zero, as the example in (i) realizes.
\textbf{(iii)}
With $\epsilon=0$, $\max(r_{\max}^{(t)},0)=r_{\max}^{(t)}$ for $\mathbf{r}^{(t)}\ge\mathbf{0}$,
i.e., the max-normalized temperature. \qed

\subsection{Proof of Proposition~\ref{prop:hr}}
\label{app:proof-hr}

\textbf{(i)} Directly from Eqn.~\eqref{eq:hr}.
\textbf{(ii)}
Primes denote contaminated quantities; $L$ and $U$ are the smallest and largest other observations in
$[s_0-H,s_0+H]$.
As the median uses raw observations, $\mu_d'^{(s)}\neq\mu_d^{(s)}$ only if $s\in(s_0,s_0+H]$,
and for $s>s_0+H$ the window and current observation are clean, so $\tilde m_d'^{(s)}=\tilde m_d^{(s)}$.
For $s=s_0$, the window $\{s_0-H,\dots,s_0-1\}$ is clean (as $s_0>H$) and lies in $[s_0-H,s_0+H]$,
so $\mu_d'^{(s_0)}\in[L,U]$, and (i) gives $\tilde m_d'^{(s_0)}\in[e^{-c}L,e^{c}U]$ for any $M$.
For $s\in(s_0,s_0+H]$, the window contains the contaminated value and $H-1\ge 2$ clean values in $[L,U]$.
The median involves only ranks in $[2,H-1]$ (as $H\ge 3$): the $k$-th order statistic with $k=(H+1)/2$ for odd
$H$, or the mean of the $k$-th and $(k+1)$-th with $k=H/2$ for even $H$.
Each such order statistic lies between two order statistics of the remaining $H-1$ values,
so $\mu_d'^{(s)}\in[L,U]$.
The clean $m_d^{(s)}\in[L,U]$, and clipping moves it toward $\mu_d'^{(s)}$, so $\tilde m_d'^{(s)}\in[L,U]\subseteq[e^{-c}L,e^{c}U]$.
An EMA level is a convex combination of clipped observations, each equal to its clean counterpart or in
$[e^{-c}L,e^{c}U]$, so it lies in an interval independent of $M$; without clipping, the EMA weight $\omega_i>0$
of step $s_0$ makes it at least $\omega_iM$.

\textbf{(iii)}
Let $m_d^{(s)}=A$ for $s<s_0$ and $m_d^{(s)}=A'$ for $s\ge s_0$.
For $s\ge s_0+\lfloor H/2\rfloor+1$, the window of $\mu_d^{(s)}$ (of size at most $H$) contains at least
$\lfloor H/2\rfloor+1$ values equal to $A'$ (a strict majority),
so $\mu_d^{(s)}=A'=m_d^{(s)}$ lies in the clipping range and $\tilde m_d^{(s)}=m_d^{(s)}$. \qed

\subsection{Proof of Theorem~\ref{thm:tilt}}
\label{app:proof-tilt}
The full statement adds:
in (i), the maximizer is the KL projection of $\mathbf{p}\odot e^{\eta\mathbf{h}}$ (normalized) onto $\mathcal{U}$;
(ii) also gives $q_d\ge\max(\lambda_d,\kappa^{-\eta}p_d)$ and $\mathrm{KL}(\mathbf{q}\,\|\,\mathbf{p})\le\eta\log\kappa$;
in (i) with $\rho=\mathrm{KL}(\mathbf{q}\,\|\,\mathbf{p})$, when the floor is inactive,
$\sup_{\mathrm{KL}(\mathbf{q}'\|\mathbf{p})\le\rho}\langle\mathbf{q}',\mathbf{h}\rangle
=\inf_{\gamma>0}\big\{\gamma\rho+\gamma\log\mathbb{E}_{d\sim\mathbf{p}}[e^{h_d/\gamma}]\big\}$,
attained at $\gamma=1/\eta$.
Let $Z=\langle\mathbf{p},e^{\eta\mathbf{h}}\rangle\ge 1$ (as $h_d\ge 0$) and $\mathbf{w}=\mathbf{p}\odot e^{\eta\mathbf{h}}/Z\in\Delta^{K}$.

\textbf{(i), unique maximizer.}
For any $\mathbf{q}\in\Delta^{K}$, expanding $\mathrm{KL}(\mathbf{q}\,\|\,\mathbf{w})$ gives
\begin{equation}
  \langle\mathbf{q},\mathbf{h}\rangle-\tfrac{1}{\eta}\mathrm{KL}(\mathbf{q}\,\|\,\mathbf{p})
  =\tfrac{1}{\eta}\log Z-\tfrac{1}{\eta}\mathrm{KL}(\mathbf{q}\,\|\,\mathbf{w}).
  \label{eq:proof-gibbs}
\end{equation}
Hence Eqn.~\eqref{eq:rds-dro} is the KL projection $\min_{\mathbf{q}\in\mathcal{U}}\mathrm{KL}(\mathbf{q}\,\|\,\mathbf{w})$
onto the convex $\mathcal{U}$, unique by strict convexity.
With multipliers $\chi$ and $\beta_d$ of the simplex and floor constraints, the KKT conditions
$\log(q_d/w_d)+1+\chi-\beta_d=0$, $\beta_d\ge 0$, and $\beta_d(q_d-\lambda_d)=0$ give $q_d=a\,w_d$ if $q_d>\lambda_d$, and $q_d=\lambda_d\ge a\,w_d$ otherwise, with $a=e^{-1-\chi}$,
i.e., $q_d=\max(\lambda_d,a\,w_d)$, which is Eqn.~\eqref{eq:tilt} with $\nu=a/Z$.
$G(a)=\sum_d\max(\lambda_d,a\,w_d)$ is continuous and nondecreasing with $G(0)=\sum_d\lambda_d<1\le G(1)$,
so $G(a)=1$ has a root in $(0,1]$, and it is unique: at any root, some $d$ has $a\,w_d\ge\lambda_d$
(otherwise $G(a)=\sum_d\lambda_d<1$), and as $w_d>0$, $G(a')\ge G(a)+(a'-a)w_d>1$ for all $a'>a$.

\textit{Exactness of the implementation.}
From $\mathbf{w}$ (Appendix~\ref{app:sched}), the unclamped coordinates after round $k$ are $b_k w_d$, $b_0=1$.
Clamping only raises coordinates, so $b_{k+1}<b_k$; the clamped set never shrinks, and the unclamped set is never
empty as $\sum_d\lambda_d<1$, so the procedure stops within $K$ rounds, where every unclamped $d$ satisfies $b_kw_d\ge\lambda_d$, and every clamped $d$, clamped at some round
$j\le k$, satisfies $b_kw_d\le b_jw_d<\lambda_d$.
Hence $\mathbf{q}=\max(\boldsymbol{\lambda},b_k\mathbf{w})$ with a unit sum, and $b_k=a$ by uniqueness.

\textbf{(i), KL ball.}
For any $\mathbf{q}'\in\mathcal{U}$ with $\mathrm{KL}(\mathbf{q}'\,\|\,\mathbf{p})\le\rho$,
the optimality of $\mathbf{q}$ in Eqn.~\eqref{eq:rds-dro} gives
$\langle\mathbf{q}',\mathbf{h}\rangle\le\langle\mathbf{q},\mathbf{h}\rangle
+\frac{1}{\eta}\big(\mathrm{KL}(\mathbf{q}'\,\|\,\mathbf{p})-\rho\big)\le\langle\mathbf{q},\mathbf{h}\rangle$,
and $\mathbf{q}$ is feasible, so it solves the constrained problem.
For the dual form, Eqn.~\eqref{eq:proof-gibbs} with $1/\gamma$ in place of $\eta$ gives,
for any $\gamma>0$ and $\mathrm{KL}(\mathbf{q}'\,\|\,\mathbf{p})\le\rho$,
$\langle\mathbf{q}',\mathbf{h}\rangle\le\gamma\,\mathrm{KL}(\mathbf{q}'\,\|\,\mathbf{p})
+\gamma\log\mathbb{E}_{\mathbf{p}}[e^{h_d/\gamma}]\le\gamma\rho+\gamma\log\mathbb{E}_{\mathbf{p}}[e^{h_d/\gamma}]$,
i.e., weak duality~\citep{DBLP:journals/mor/Lam16}.
When the floor is inactive, $\mathbf{q}=\mathbf{w}$ and Eqn.~\eqref{eq:proof-gibbs} gives
$\langle\mathbf{q},\mathbf{h}\rangle=\frac{1}{\eta}\rho+\frac{1}{\eta}\log Z$, i.e., equality at $\gamma=1/\eta$.

\textbf{(ii), bounded deviation.}
Let $F=\{d:a\,w_d\ge\lambda_d\}$ and $C$ its complement; $F\neq\emptyset$ as $\sum_d\lambda_d<1$.
Let $\psi_d=a\,e^{\eta h_d}/Z$, so that $q_d/p_d=\psi_d$ for $d\in F$, and
$q_d/p_d=\lambda_d/p_d\in(\psi_d,1]$ for $d\in C$.
\textit{Upper bound:} for $d\in F$, $q_d/p_d=\psi_d\le e^{\eta h_d}\le\kappa^{\eta}$ as $a\le 1\le Z$;
for $d\in C$, $q_d/p_d\le 1$.
\textit{Lower bound:} if $\max_{d\in F}\psi_d<1$, all ratios are below one on $F$ and at most one on $C$, contradicting
$\sum_dp_d(q_d/p_d)=1$; hence some $d'\in F$ has $\psi_{d'}\ge 1$, and for every $d$, $q_d/p_d\ge\psi_d=\psi_{d'}e^{\eta(h_d-h_{d'})}\ge\kappa^{-\eta}$,
since $h_d-h_{d'}\ge-\log\kappa$.
Finally, $\mathrm{KL}(\mathbf{q}\,\|\,\mathbf{p})=\sum_dq_d\log(q_d/p_d)\le\max_d\log(q_d/p_d)\le\eta\log\kappa$.
\textbf{(ii), no starvation.}
$q_d\ge\lambda_d=\min(p_d,\lambda_{\mathrm{lo}})$ by feasibility, and $p_d\ge f$ by
Eqn.~\eqref{eq:rds-nominal} as $Kf\le 1$, so $q_d\ge\min(f,\lambda_{\mathrm{lo}})$.

\textbf{(iii)}
If $h_d\equiv h_0$, then $\mathbf{w}=\mathbf{p}\in\mathcal{U}$, so $\mathbf{q}=\mathbf{p}$.
Let $d^\dagger\in\arg\max_dh_d$, so that $\psi_{d^\dagger}=\max_d\psi_d$.
If $d^\dagger\in C$, then every $d\in F$ has $q_d/p_d=\psi_d\le\psi_{d^\dagger}<\lambda_{d^\dagger}/p_{d^\dagger}\le 1$,
and every $d\in C$ has $q_d/p_d\le 1$, contradicting $\sum_dp_d(q_d/p_d)=1$.
Hence $d^\dagger\in F$, and by (ii), $q_{d^\dagger}/p_{d^\dagger}=\max_{d\in F}\psi_d\ge 1$. \qed